\def\year{2022}\relax
\documentclass[letterpaper]{article}
\usepackage{aaai22} % DO NOT CHANGE THIS
\usepackage{times} % DO NOT CHANGE THIS
\usepackage{helvet} % DO NOT CHANGE THIS
\usepackage{courier} % DO NOT CHANGE THIS
\usepackage[hyphens]{url} % DO NOT CHANGE THIS
\usepackage{graphicx} % DO NOT CHANGE THIS
\usepackage{graphicx} % DO NOT CHANGE THIS
\usepackage{natbib} % DO NOT CHANGE THIS
\usepackage{caption} % DO NOT CHANGE THIS
\DeclareCaptionStyle{ruled}%
{labelfont=normalfont,labelsep=colon,strut=off}
\usepackage{algorithm}
\usepackage{algpseudocode}
\usepackage{graphicx}
\usepackage{amsmath}
\usepackage{amssymb}
\usepackage{multirow}
\usepackage{mathtools}
\usepackage{xcolor}
\usepackage{colortbl}
\usepackage{xspace}
\usepackage{xfrac}

\makeatletter
\DeclareRobustCommand\onedot{\futurelet\@let@token\@onedot}
\def\@onedot{\ifx\@let@token.\else.\null\fi\xspace}

\def\eg{\emph{e.g}\onedot} 
\def\ie{\emph{i.e}\onedot}

\makeatother

\usepackage{booktabs}       % professional-quality tables
\usepackage{nicefrac}       % compact symbols for 1/2, etc.
\usepackage{microtype}
\usepackage{multicol}
\usepackage{arydshln}
\usepackage{graphicx}
\usepackage{newfloat}
\usepackage{listings}
\floatstyle{ruled}
\newfloat{listing}{tb}{lst}{}
\floatname{listing}{Listing}
\title{Iterative Contrast-Classify For Semi-supervised Temporal Action Segmentation}
\author {
    Dipika Singhania \textsuperscript{\rm 1},
    Rahul Rahaman \textsuperscript{\rm 1},
    Angela Yao \textsuperscript{\rm 1}
}
\affiliations {
    \textsuperscript{\rm 1} National University of Singapore\\
    dipika16@comp.nus.edu.sg, 
    rahul.rahaman@u.nus.edu, 
    ayao@comp.nus.edu.sg
}
\renewcommand{\epsilon}{\varepsilon}
\renewcommand{\phi}{\varphi}

\DeclareMathOperator{\argmax}{\arg\,\max}

\newcommand\numberthis{\addtocounter{equation}{1}\tag{\theequation}}

\newcommand{\calD}{\mathcal{D}}
\newcommand{\calL}{\mathcal{L}}

\newcommand{\calA}{\mathcal{A}}

\newcommand{\calP}{\mathcal{P}}

\newcommand{\calF}{\mathcal{F}}
\newcommand{\calN}{\mathcal{N}}
\newcommand{\calI}{\mathcal{I}}

\newcommand{\bbR}{\mathbb{R}}
\newcommand{\bbI}{\mathbb{I}}
\newcommand{\bbN}{\mathbb{N}}

\newcommand{\bp}{\mathbf{p}}

\newcommand{\bG}{\mathbf{G}}
\newcommand{\bz}{\mathbf{z}}

\newcommand{\bh}{\mathbf{h}}

\newcommand{\feat}{\mathbf{f}}

\newcommand{\vid}{\ensuremath{\text{V}}}
\newcommand{\enc}{\ensuremath{\mathbf{\Phi}}}
\newcommand{\dec}{\ensuremath{\mathbf{\Psi}}}
\newcommand{\bottle}{\ensuremath{\mathbf{\Gamma}}}

\newcommand{\simi}[2]{\text{sim}\!\left( #1 , #2 \right)}

\newcommand{\losscons}{\ensuremath{\calL_{\text{con}}}}
\newcommand{\losscross}{\ensuremath{\calL_{\text{ce}}}}

\newcommand{\BK}[1]{ {\left( #1 \right)} }
\newcommand{\sqBK}[1]{ {\left[ #1 \right]} }

\newcommand{\norm}[1]{\left\Vert #1 \right\Vert}
\DeclarePairedDelimiter\ceil{\lceil}{\rceil}
\DeclarePairedDelimiter\floor{\lfloor}{\rfloor}

\definecolor{LightCyan}{rgb}{0.88,1,1}
\definecolor{LightPink}{rgb}{1,0.88,1}
\definecolor{LightGreen}{rgb}{0.88,1,0.88}
\definecolor{LightOrange}{rgb}{1,0.94,0.88}

\begin{document}

\maketitle

\begin{abstract}
Temporal action segmentation classifies the action of each frame in (long) video sequences. Due to the high cost of frame-wise labeling, we propose the first semi-supervised method for temporal action segmentation.
Our method hinges on unsupervised representation learning, which, for temporal action segmentation, poses unique challenges.  Actions in untrimmed videos vary in length and have unknown labels and start/end times. Ordering of actions across videos may also vary.  We propose a novel way to learn frame-wise representations from temporal convolutional networks (TCNs) by clustering input features with added \textit{time-proximity condition} and \textit{multi-resolution similarity}. 
 By merging representation learning with conventional supervised learning, we develop an \textit{``Iterative-Contrast-Classify (ICC)''} semi-supervised learning scheme. With more labelled data, ICC progressively improves in performance; ICC semi-supervised learning, with 40\% labelled videos, performs similar to fully-supervised counterparts. Our ICC improves MoF by \{+1.8, +5.6, +2.5\}\% on Breakfast, 50Salads and GTEA respectively for 100\% labelled videos.

\end{abstract}

\section{Introduction}\label{sec:introduction}
Temporal action segmentation takes long untrimmed video containing multiple actions in a sequence and estimates the action labels for each video frame. 
There is a huge annotation cost to label each frame of all videos for action segmentation, especially as the videos are minutes long.
Several works aim to reduce annotation requirements 
with weak supervision like transcripts~\citep{weakly-chang2019d3tw}, 
or few frame labels~\cite{timestamp-weakly-li2021temporal}.   
In this work, we advocate using semi-supervised learning, \ie having labels only for a fraction of the videos in the training set. Specifically, we design the unsupervised representation learning step to learn the underlying distribution of all unlabelled videos, which helps achieve higher temporal action segmentation scores with very few labeled videos in the subsequent supervised training step.

\begin{figure}[t]
\centering
\includegraphics[width=0.9\linewidth]{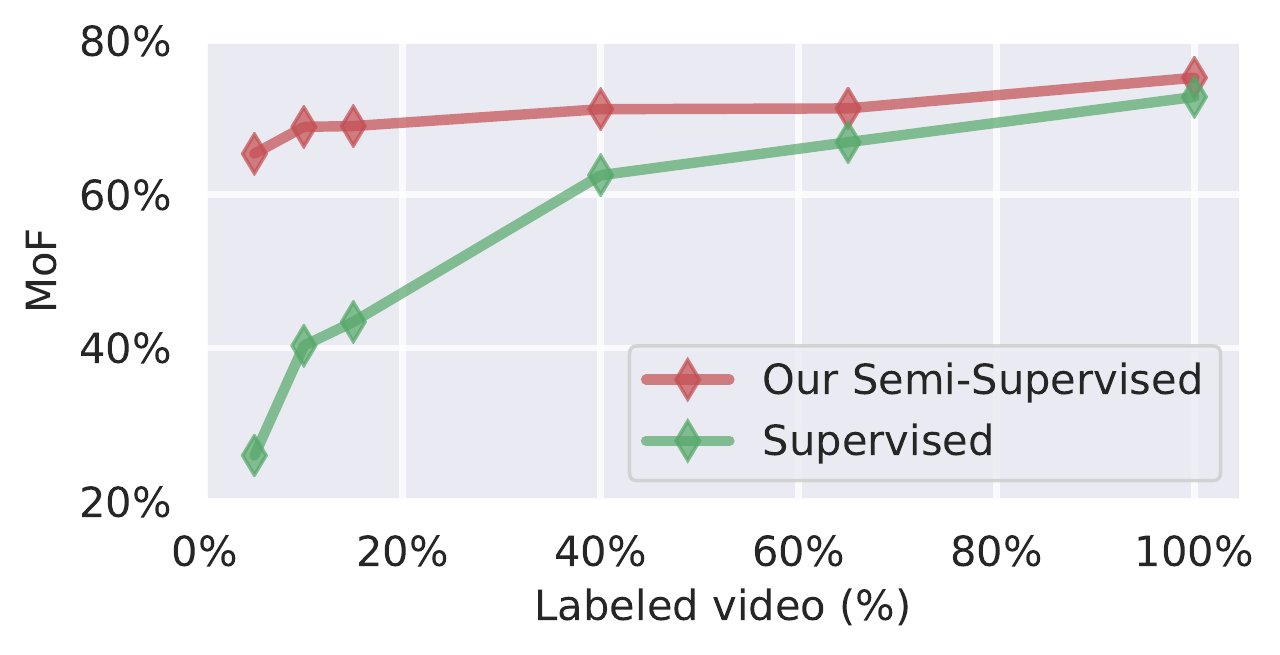}
\caption{\textbf{Frame accuracy on Breakfast dataset:} Our semi-supervised approach has impressive performance with just 5\% labelled videos; at 40\%, we almost match the Mean over Frames (MoF) of a 100\% fully-supervised setup. 
}
\label{fig:teaser}
\end{figure}

\begin{figure*}[t!]
    \centering
    \includegraphics[width=0.9\linewidth]{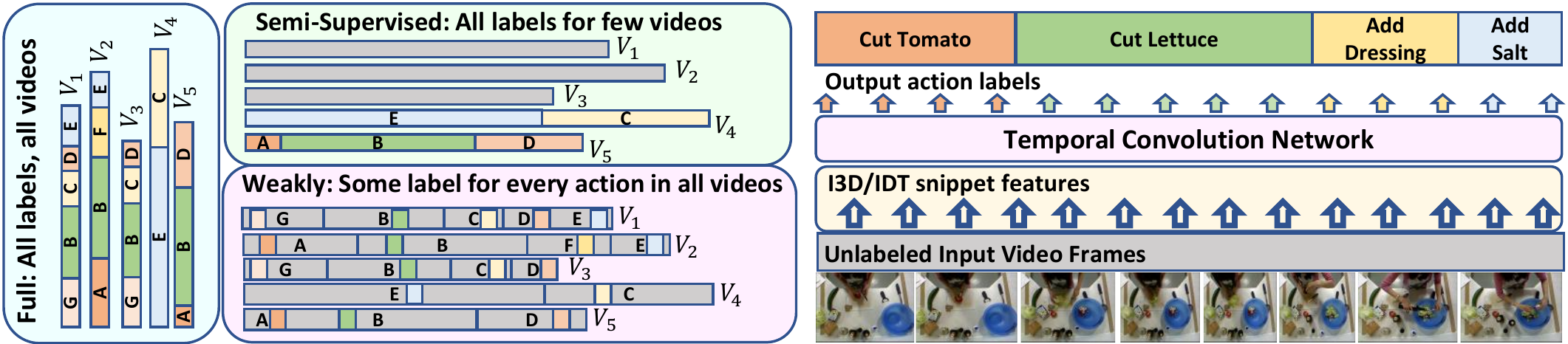}
    \caption{Left: Comparison on forms of supervision. Right: Overview of Temporal Action Segmentation task with TCN.}
    \label{fig:supervision_form}
\end{figure*}

For unsupervised representation learning, we are inspired by the success of contrastive learning in images~\cite{simCLR}, short-trimmed videos ~\cite{time-contrast-3-lorre2020temporal, temporal-contrast-singh2021semi} and
other areas of machine learning ~\citep{constrast3-chen2021momentum, contrast6-rahaman2021pretrained}. 
Works which apply contrastive learning to longer sequences bring together multiple viewpoints of a sequence~\cite{time-contrast-2-sermanet2018time} or multiple modalities, such as video and text~\cite{time-contrast-4-alwassel2019self} or video and audio~\cite{time-contrast-6-miech2020end}. These settings target multi-view or multi-modal representations and are not applicable for videos in action segmentation datasets. 
Also, standard contrastive technique to bring image (or video) and its augmentations near is unlikely to be effective. Action segmentation is a frame-wise (and not video-wise) classification task so a model should capture similarities across temporally disjoint but semantically similar frames, while factoring temporal continuity within every action segment.  The latter is easy to incorporate in the form of temporal constraints, but the former poses significant challenges without action labels. 

As such, contrastive learning has not yet been explored for action segmentation and our work is the first. We design a novel strategy to form the positive and negative sets without labels by leveraging the discriminativeness of the pre-trained I3D~\cite{carreira2017quo} input features (see Fig.~\ref{fig:supervision_form} right). 
 As a base model, we use a \textit{SOTA} temporal convolutional network (TCN)~\cite{singhania2021coarse}; a key advantage is the progressive temporal upsampling in the decoder which allows us to integrate contrastive learning to multiple temporal resolutions and enforce temporal continuity by design. Our proposed multi-resolution representation for contrastive learning is previously unexplored and 
is highly effective for temporal action segmentation.

Equipped with our (unsupervised) learned model, we can perform semi-supervised learning with only a small fraction of labelled training videos. To fully utilize the labeled and unlabeled dataset, we propose a novel \textit{Iterative-Contrast-Classify} algorithm that 
updates the representations while learning to segment sequences and assigning pseudo-labels to the unlabelled videos. 
We achieve noteworthy segmentation performance with just 5\% labelled videos, while with 40\% labels, we can almost match the full-supervision (see Fig.~\ref{fig:teaser}).

To the best of our knowledge, our work is the first to apply semi-supervised learning for temporal action segmentation. The closest in spirit are SSTDA~\cite{selfsupervised-chen2020action}, TSS\cite{timestamp-weakly-li2021temporal}. However, these works are weakly-supervised setups and require (weak) labels for \textit{every} training video (see Fig.~\ref{fig:supervision_form} left).  TSS require one frame label for each action\footnote{Our setup is analogous to semi-supervised image segmentation~\cite{adversarialSSL2DSemanticSeg,ConsistSSL2DSemanticSeg}: most training images are un-annotated, while a few are fully-annotated. The analogue of TSS is point-supervision~\cite{pointsupervision-bearman2016s}, \ie labelling one pixel from each object of every training image.}  While the percentage of overall labeled frames is very little (0.03\%), the annotation effort should not be under-estimated.  Annotators must still watch \textit{all} the videos 
and labelling timestamp frames gives only a 6X speedup compared to densely labelling all frames~\cite{singleframe-sfnet-ma2020sf}.

Summarizing our contributions, we:
\begin{itemize}
    \item Proposed a novel unsupervised representation learning that leverages the discriminativeness in pre-trained input features and temporal continuity in a video sequence.
    \item Designed a multi-resolution representation for contrastive learning which inherently encodes sequence variations and temporal continuity. 
    \item Formulated a new semi-supervised learning variant of temporal action segmentation and proposed an \emph{``Iterative-Contrast-Classify (ICC)''} algorithm that iteratively fine-tunes representations and strengthens segmentation performance with few labelled videos.
\end{itemize}

\section{Related Work} \label{sec:related_work}

\subsubsection{Temporal action segmentation} Classifying and temporally segmenting fine grained actions in long video requires both local motion and global long-range dependencies information. It is standard to extract snippet level IDT~\cite{wang2013action} or  
{I3D} features 
to capture local temporal motion (and to reduce computational expense of joint end-to-end training). These features are used as inputs to Temporal Convolution Networks (TCNs) 
which captures global action compositions, segment durations and long-range dependencies (see Fig.\ref{fig:supervision_form} right).
Fully supervised frameworks require per-frame annotations of all the video sequences in the dataset. Popular TCN frameworks include 
U-Net style encoder-decoders~\citep{TED-lea2017temporal, TEDresi-lei2018temporal, singhania2021coarse} or 
temporal resolution preserving MSTCNs~\citep{li2020ms, farha2019ms}. 

Weakly supervised methods bypass per-frame annotations and use labels such as ordered lists of actions ~\citep{TED-ding2018weakly, weakly-richard2018neuralnetwork, weakly-chang2019d3tw, weakly-li2019weakly, weakly-souri2019fast}
or a small percentage of action time-stamps~\citep{timestamp-weakly-kuehne2018hybrid,  timestamp-weakly-li2021temporal, selfsupervised-chen2020action} for \textit{all} videos. 

Unsupervised approaches use clustering, including $k$-means~\citep{unsupervised-kukleva2019unsupervised}, agglomerative~\citep{sarfraz2021temporally}, and discriminative clustering~\citep{unsupervised-sener2018unsupervised}. To improve clustering performance, some works~\cite{unsupervised-kukleva2019unsupervised, unsupervised-vidalmata2021joint} learns representation by predicting frame-wise feature's absolute temporal positions in the video.
Our unsupervised representation implicitly capture relative temporal relationships based on temporal distance rather than absolute positions. 

\subsubsection{Unsupervised Contrastive Feature Learning} dates back to~\cite{contrast6-hadsell2006dimensionality} but was more recently formalized in SimCLR~\cite{simCLR}. Most works~\citep{constrast3-chen2021momentum, contrast6-rahaman2021pretrained, contrast4-he2020momentum, contrast5-khosla2020supervised} hinge on well-defined data augmentations, with the goal of bringing together the original and augmented sample in the feature space. 

The few direct extensions of SimCLR for video~\citep{time-contrast-5-bai2020can, time-contrast-7-qian2020spatiotemporal, time-contrast-3-lorre2020temporal} target action recognition on few seconds short clips. Others integrate contrastive learning 
by bringing together next-frame feature predictions with actual representations~\cite{time-contrast-1-kong2020cycle, time-contrast-3-lorre2020temporal}, using path-object tracks for bringing cycle-consistency~\cite{temporal-constrast-wang2020contrastive}, and considering multiple viewpoints~\cite{time-contrast-2-sermanet2018time} or accompanying modalities like audio~\cite{time-contrast-4-alwassel2019self} or text~\cite{time-contrast-6-miech2020end}.  We are inspired by these works to develop contrastive learning for long-range segmentation. However, previous works differ fundamentally in both aim~\ie learning the underlying distribution of cycle-consistency in short clips, and input data~\eg multiple viewpoints or modalities.

\begin{figure*}[t!]
\centering
  \includegraphics[width=0.85\textwidth]{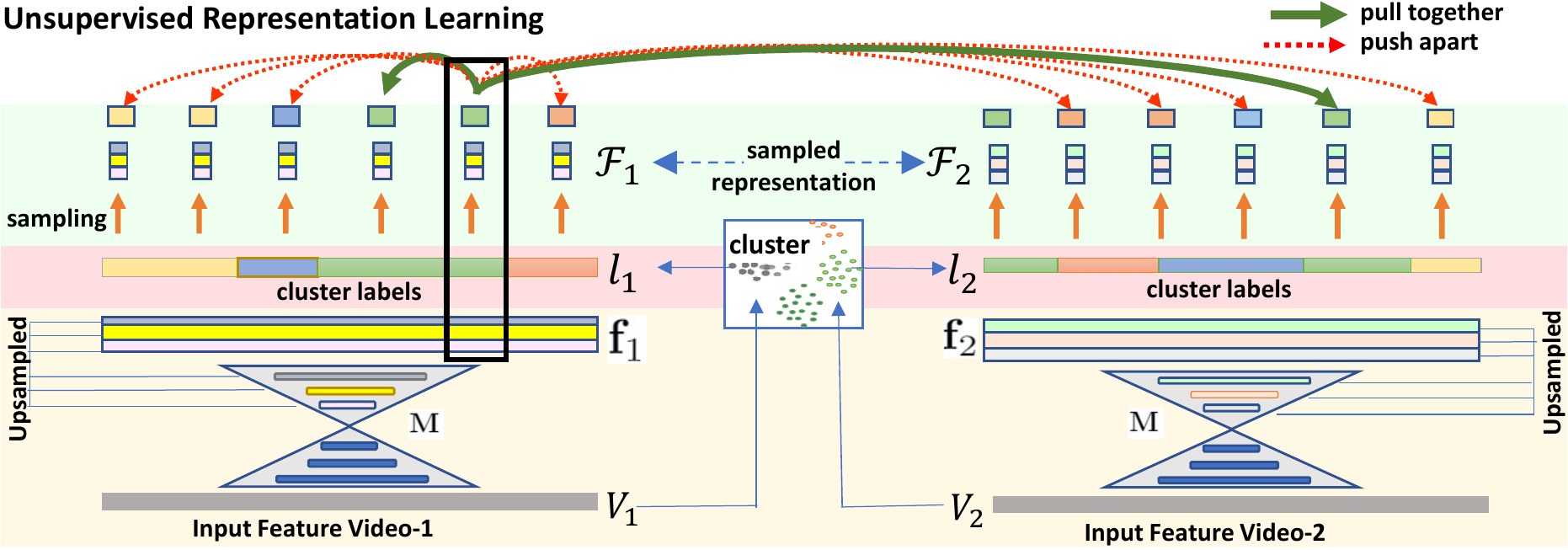}
  \caption{\textbf{Unsupervised Representation Learning.} Step 1 (bottom orange panel): Pass pre-trained I3D inputs $V$ into the base TCN and generate multi-resolution representation $\feat$. Step 2 (middle pink panel): Cluster the I3D inputs $V$ within a training mini-batch and generate frame-wise cluster labels $l$. Step 3 (top green panel): Representations $\feat$ and its corresponding cluster label $l$ is sampled based on temporal proximity sampling strategy to form feature set $\mathcal{F}$. Step 4: Apply contrastive learning to ``pull together'' (green arrows) similar samples in the positive set and ``push apart'' (red arrows) other samples in the negative set.
 } 
 \label{fig:contrastive_pic}
\end{figure*}

\section{Preliminaries} \label{sec:preliminaries}
\subsubsection{Definitions:} \label{subsec:definitions} 
We denote a video as $\text{V} \in \bbR^{T \times F}$; for each temporal location $t\!<\!T$, 
frame $\vid[t] \in \bbR^F$ is a $F$-dimensional pre-trained I3D feature.
Note, input I3D feature is from model pre-trained on Kinetics dataset ~\cite{carreira2017quo} and is not fine-tuned on our segmentation datasets. 

For simplicity, unless otherwise explicitly noted, \eg in Section~\ref{subsec:multi-resolution-similarity}, we treat the temporal dimension of all the videos as a normalized unit interval $t \in [0,1]$,~\ie $T\!=\!1$. Each frame $\vid[t]$ has a ground truth action label $y[t] \in 
\{1,...,A\}$ from a pre-defined set of $A$ action classes.  Additionally, each video has a higher-level \textit{complex activity} label $c \in \{1,\ldots,C\}$. The complex activity specifies an underlying objective, \eg \emph{`making coffee'} for action sequence $\{$\emph{`take cup'}, \emph{`pour coffee'}, \emph{`add sugar'}, \emph{`stir'}$\}$, though the sequence ordering may differ \eg \emph{`add sugar'} may come before \emph{`pour coffee'}.

\subsubsection{Base Segmentation Model:}\label{subsec:base_segmentation}
We use 
the C2F-TCN~\citep{singhania2021coarse}, which is a U-Net style encoder-decoder, though our method is also applicable to other base models such as the ED-TCN~\citep{TED-lea2017temporal}. The encoder layers \enc{} take pre-trained video features as inputs and progressively increases the feature abstraction while reducing the temporal resolution up to some bottleneck \bottle. The decoder layers \dec{} then increase the temporal resolution symmetrically with respect to the encoder.  We refer to the Appendix-B.1 or to~\cite{singhania2021coarse} for more details.  The overall encoder-decoder $\mathbf{M}\!:=\!\BK{\enc:\bottle:\dec}$ takes $\vid$ as input and produces output frame-wise features $\feat = \mathbf{M}(\vid)$.  For each time $t, \feat[t] \in \bbR^d$ is a $d$-dimensional representation.  We describe in detail how $\feat$ is formed in Section~\ref{subsec:multi-resolution-similarity}.

\subsubsection{Learning Framework \& Data Split:}\label{subsec:framework_splits}
Our overall framework has two stages.  First, we apply an unsupervised representation learning to learn a model $\mathbf{M}$ (Section~\ref{subsec:unsupervised_representation_learning}). Subsequently, model $\mathbf{M}$ is trained with linear projection layers (action classifiers) with a small portion of the labelled training data to produce the semi-supervised model ($\mathbf{M}:\mathbf{G}$)
(Section~\ref{sec:semi-supervised-temporal-segmentation}). For representation learning, we follow the convention of previous unsupervised works ~\cite{unsupervised-kukleva2019unsupervised, unsupervised-vidalmata2021joint} 
in which actions $y$ are unknown, but the complex activity of each video \emph{is} known\footnote{The label is used implicitly, as the unsupervised methods are applied to videos of each complex activity individually.}.  For the semi-supervised stage, we use the ground truth $y$ for a small subset of labelled video $\calD_L$ out of a larger training dataset $\calD = \calD_U \cup \calD_L$, where $\calD_U$ denotes the unlabelled videos.

\subsubsection{Contrastive Learning}
We use contrastive learning for our unsupervised frame-wise representation learning. Following the formalism of~\cite{simCLR}, we define a set of features $\calF\!:=\!\{\feat_i, i\!\in\!\calI\}$ indexed by a set $\calI$. Each feature $\feat_i\!\in\!\calF$ is associated with two disjoint sets of indices $\calP_i\!\subset\!\calI\!\setminus\!\{i\} $ and $\calN_i\!\subset\!\calI\!\setminus\!\{i\}$. The features in the positive set $\calP_i$ should be similar to $\feat_i$, while the features in the negative set  $\calN_i$ should be contrasted with $\feat_i$. For each $j \in \calP_i$, the contrastive probability $p_{ij}$ is defined as

\begin{equation}\label{eqn:contrastive_prob}
    p_{ij} = \frac{e_{\tau}\BK{\feat_i, \feat_j}}{e_{\tau}\BK{\feat_i, \feat_j} + \sum_{k \in \calN_i} e_{\tau}\BK{\feat_i, \feat_k}},
\end{equation}

\noindent where the term $e_{\tau}=\exp\{\text{cos}(\feat_i,\feat_j)/\tau\} $ is the exponential of the cosine similarity between $\feat_i$ and $\feat_j$ scaled by temperature $\tau$. Maximizing the probability in Eq.~\eqref{eqn:contrastive_prob} ensures that $\feat_i, \feat_j$ are similar while also decreasing the cosine similarity between $\feat_i$ and any feature in the negative set.  The key to effective contrastive learning is to identify the relevant positive and negative sets to perform the targeted task.

\section{Unsupervised Representation Learning}\label{subsec:unsupervised_representation_learning}
Our representation learning applies contrastive learning at frame-level, based on input feature clustering and temporal continuity (Sec.~\ref{sec:frame_level_contratsive}), and at a video-level, by leveraging the complex activity labels (Sec.~\ref{subsec:video-level-contrast}).  We merge the two objectives into a common loss that is applied to our multi-temporal resolution feature representations (Sec.~\ref{subsec:multi-resolution-similarity}). 

\subsection{Frame-Level Contrastive Formulation}\label{sec:frame_level_contratsive}

\subsubsection{Input Clustering:}\label{subsubsec:input_feature_cluster}
Our construction of positive and negative sets should respect the distinction between different action classes.  But as our setting is unsupervised, there are no labels to guide the formation of these sets. Hence we propose to leverage the discriminative properties of the pre-trained input I3D features to initialize the positive and negative sets.  Note that while the clusters are formed on the input features, our contrastive learning is done over the representation $\feat$ produced by the C2F-TCN model (yellow panel of Fig.~\ref{fig:contrastive_pic}). 

Specifically, we cluster the individual frame-wise inputs $\vid[t]$ for all the videos within a small batch. We use k-means clustering and set the number of clusters as $2A$ (ablations in Appendix-A.1), \ie{} twice the number of actions to allow variability even within the same action. After clustering, each frame $t$ is assigned the cluster label $l[t] \in \{1,\ldots,2A\}$.  Note 
that this simple clustering does not require videos of the same (or different) complex activities to appear in a mini-batch.  It also does not incorporate temporal information -- this differs from previous unsupervised works~\cite{unsupervised-kukleva2019unsupervised, unsupervised-vidalmata2021joint} that embed absolute temporal locations into the input features \textit{before} clustering.  

\subsubsection{Representation Sampling Strategy:} \label{sec:representation_sampling}  
The videos used for action segmentation are long, \ie 1-18k frames. Contrasting all the frames of every video in a batch would be too computationally expensive to consider, whereas contrastive loss of even a few representation back-propagates through the entire hierarchical TCN. To this end, we dynamically sample a fixed number of frames from each video to form the feature (representation) set $\calF$ for each batch of videos. Note that the sampling is applied to the feature representations $\feat=\mathbf{M}(\vid)$ and not to the inputs $\vid$; and the full input $\vid$ is required to pass through the TCN to generate $\feat$. 

Let $\calI$ denote the feature set index (as in sec \ref{sec:preliminaries}) and for any feature index $(n,i) \!\in\! \calI$, let $n$ denote the video-id and $i$ the sample-id within that video. For a video $\vid_n$ and a fixed $K\!>\!0$, we sample $2K$ frames $\{t^n_i: i\!\le\! 2K\} \subset [0,1]$ and obtain the feature set $\calF_n\!:=\!\{\feat_n[t^n_i]\!:\!i\!\le\!2K\}$.  To do so, we divide the unit interval $[0,1]$ into $K$ equal partitions 
and randomly choose a single frame from each partition. Another $K$ frames are then randomly chosen %a distance of $1/(2K)$ %
$\epsilon$ away ($\epsilon \ll 1/K$)
from each of the first $K$ samples. This strategy ensures diversity %across the video 
(the first $K$ samples) while having nearby $\epsilon-$distanced features (the second $K$ samples) to either enforce temporal continuity if they are the same action, or learn boundaries if they are different actions (approximated by the cluster labels $l$ when actions labels are unknown). 

\subsubsection{\textbf{Frame-level positive and negative sets:}}\label{subsubsec:positive-negative-sets} Constructing the positive and negative set for each index $(n,i) \in \calI$ requires a notion of similar features. The complex activity label is a strong cue, as there are either few or no shared actions across the different complex activities.  For video $\vid_n$ with complex activity $c_n$, we contrast index $(m, j)$ with $(n, i)$ if $c_m \neq c_n$. In datasets without meaningful complex activities (50Salads, GTEA), this condition is not applicable. 

The cluster labels $l$ of the input features
already provides some separation between actions (see Table \ref{tab:unsupervised_ablation}); we impose an additional temporal proximity condition to minimize the possibility of different action in the same cluster. Formally, we bring the representation with index $(n,i)$ close to $(m,j)$ if their cluster labels are same i.e $l_n[t^n_i] = l_m[t^m_j]$ \emph{and} if they are close-by in time, i.e.,~$|t^n_i - t^m_j| \leq \delta$. For datasets with significant variations on the action sequence, \eg 50Salads, the same action may occur at very different parts of the video  
so we choose higher $\delta$, vs. smaller $\delta$ for actions that follow more regular ordering, \eg Breakfast.
Sampled features belonging to the same cluster but exceeding the temporal proximity, \ie $l_n[t^n_i] = l_m[t^m_j]$ but $|t^n_i - t^m_j| > \delta$ are not considered for neither the positive nor the negative set.

Putting together the criteria from complex activity labels, clustering and temporal proximity, our positive and negative sets for index $(n,i)$, \ie{} sample $i$ from video $n$ is defined as
\begin{align*}
    \calP_{n,i}\!&=\!\{(m, j)\!: c_m=c_n, |t^n_i - t^m_j| < \delta,\, l_n[t^n_i] = l_m[t^m_j]\} \\
    \calN_{n,i} &= \{(m, j)\!: c_m \neq c_n\} \,\cup \numberthis{} \label{eqn:positive-negative}\\
    &\qquad \{(m,j)\!: c_m=c_n, l_n[t^n_i] \neq l_m[t^m_j]\}
\end{align*}
where $m, n$ are video indices, $t^n_i$ is the frame-id corresponding to the $i^{th}$ sample of video $n$, 
$c_n$ is the complex activity of video $n$, and $l_n[t^n_i]$ the cluster label of frame $t^n_i$.  
For an index $(m, j) \in \calP_{n,i}$, \ie{} belonging to the positive set of $(n,i)$, the contrastive probability becomes
\begin{gather}\label{eqn:frame-level-contrast}
    \!\!\!\! p^{nm}_{ij}\!=\!\frac{e_{\tau}\Big(\feat_n[t^n_i], \feat_m[t^m_j]\Big)}{e_{\tau}\Big( \feat_n[t^n_i], \feat_m[t^m_j] \Big) +\!\!\sum\limits_{(r,k) \in \calN_{n,i}}\!\!\!\!  e_{\tau} \Big( \feat_n[t^n_i], \feat_r[t^r_k]\Big)}.
\end{gather}
where $e_{\tau}$ is the $\tau$-scaled exponential cosine similarity of Eq.~\eqref{eqn:contrastive_prob}. For feature representations $\feat_n[t^n_i]$, Fig.~\ref{fig:contrastive_pic} visualizes the positive set with pull-together green-arrows and negative set with push apart red-arrows. 

\subsection{Video-Level Contrastive Formulation} \label{subsec:video-level-contrast}
To further emphasize global differences between different complex activities, we construct video-level summary features $\bh_n \in \bbR^d$ 
by max-pooling the frame-level features $\feat_n \in \bbR^{T_n\times d}$ along the temporal dimension.  For video $\vid_n$, we define video-level feature $\bh_n = \max_{1 \le t \le T_n} \feat_n[t]$. Intuitively, the max-pooling captures permutation-invariant features and has been effective for aggregating video segments~\cite{sener2020temporal}. 
With features $\bh_n$, we formulate a video-level contrastive learning. Reusing the index set as video-ids, $\calI = \{1,...,|\calD|\}$, we define a feature set $\calF := \{\bh_n\!: n \le |\calD|\}$, where for each video $n$, there is a positive set $\calP_n := \{m\!: c_m = c_n\}$ and a negative set as $\calN_n = \calI \setminus \calP_n$. For video $n$ and another video $m \in \calP_n$ in its positive set, the contrastive probability can be defined as
\begin{gather}\label{eqn:video-level-contrast}
    p_{nm} = \frac{e_{\tau}\BK{\bh_n, \bh_m}}{e_{\tau}\BK{\bh_n, \bh_m} + \sum_{r \in \calN_n} e_{\tau}\BK{\bh_n, \bh_r}}.
\end{gather}

For our final \textbf{unsupervised representation learning} we use \textbf{contrastive loss} function $\losscons$ that sums the video-level and frame-level contrastive losses: 
\begin{gather}\label{eqn:contrastice-loss}
    \!\!\losscons\! =\! - \tfrac{1}{N_1} \sum_n\!\!
    \sum_{m\in\!\calP_n} \log p_{nm}\! -\! \tfrac{1}{N_2}\!\! \sum_{n,i} \sum_{{m,j} \in \calP_{n,i}} \!\!\!\! \log p^{nm}_{ij},
\end{gather}
where $N_1 = \sum_n |\calP_n|, N_2 = \sum_{n,i} |\calP_{n, i}|$, and $p^{nm}_{ij}, p_{nm}$ are as defined in equation \eqref{eqn:frame-level-contrast} and \eqref{eqn:video-level-contrast} respectively. In practice, we compute this loss over mini-batches of videos.

\subsection{Multi-Resolution Representation}\label{subsec:multi-resolution-similarity}
We in this work show that constructing an appropriate representation can boost the performance of contrastive learning significantly.
For this subsection, we switch to an absolute integer temporal index, \ie for a video $\vid$ the frame indices are $t \in \{1,\ldots, T\}$ where $T \ge 1$. The decoder $\dec{}$
has six layers; each layer producing features $\bz_u, 1\!\le\!u\!\le\!6$ while progressively doubling the temporal resolution, \ie the length of $\bz_{u}$ is  $\ceil*{T/2^{6-u}}$. 
The temporally coarser features provide more global sequence-level information while the temporally fine-grained features contain more local information.  

To leverage the full range of resolutions, we combine $\{\bz_1,\ldots,\bz_6\}$ into a new feature $\bf$.  Specifically, we upsample each decoder feature $\bz_u$ to ${\hat{\bz}}_u := \text{up}\BK{\bz_u, T}$ having a common length $T$ using a temporal up-sampling function  $\text{up}\BK{\cdot, T}$ such as \textit{`nearest'} or \textit{`linear'} interpolation. 
The final frame-level representation for frame $t$ is defined as $\feat[t] = \BK{\bar{\bz}_1[t]:\bar{\bz}_2[t]:\ldots:\bar{\bz}_6[t]}$, where $\bar{\bz}_u[t] = {\hat{\bz}}_u[t]/\norm{{\hat{\bz}}_u[t]}$, ~\ie ${\hat{\bz}}_u[t]$ is normalized and then concatenated along the latent dimension for each $t$ (see Fig.~\ref{fig:contrastive_pic}). 
It immediately follows that for frames $1\!\le\! s,t\! \le\! T$, the cosine similarity $\text{cos}(\cdot)$ can be expressed as
\begin{gather}\label{eqn:similarity-linear-comb}
    \text{cos}(\feat[t],\feat[s]) = \sum_{u=1}^6 \omega_u \cdot \text{cos}(\bz_u[t],\bz_u[s]).
\end{gather}
As a result of our construction,  
the weights in Eq. \ref{eqn:similarity-linear-comb} becomes $\omega_u = \frac{1}{6}$, \ie each decoder layer an equal contribution in the cosine-similarity. Normalizing after concatenation, would cause Eq.~\eqref{eqn:similarity-linear-comb}'s coefficients $\omega_u \propto \norm{\bz_u[t]}\cdot \norm{\bz_u[s]}$. 
The importance of this ordering is verified in Appendix-C.

\subsubsection{Advantages:}\label{subsubsec:advantages} 
Our representation $\feat$ encodes some degree of temporal continuity implicitly by design. For example, in the case of `\textit{nearest}' up-sampling, it can be shown that for frames $1\!\le\! s,t\! \le\! T$, if $\floor*{t/2^u} = \floor*{s/2^u}$ for some integer $u$, then it implies that $\simi{\feat[t]}{\feat[s]} \ge 1 - u/3$ (detailed derivation in Appendix-C). Including temporally coarse features like $\bz_1$ and $\bz_2$ allows the finer-grained local features $\bz_6$ to disagree with nearby frames without harming the temporal continuity.  This makes the learned representations less prone to the common occurring problem of over-segmentation~\cite{wang2020boundary}. This is demonstrated by the significant improvement in edit, F1 scores in the last row of Table~\ref{tab:unsupervised_ablation}.

\subsection{Evaluating the Learned Representation}\label{subsec:linear-evaluation}
To evaluate the
learned representations, we train a simple linear classifier $\mathbf{G}_f$ on $\feat$ to classify frame-wise action labels. This form of evaluation is directly in line other works on unsupervised representation learning~\cite{temporal-contrast-feichtenhofer2021large, simCLR, SWAV}. The assumption is that if the unsupervised learned features are sufficiently strong, then a simple linear classifier is sufficient to separate the action classes.
Note that while our representation learning is unsupervised, learning the classifier $\mathbf{G}_f$ is fully-supervised, using ground truth labels $y$ with a cross-entropy loss $\losscross$ over the standard splits of the respective datasets. 
\begin{figure}
\centering
\includegraphics[width=0.8\linewidth]{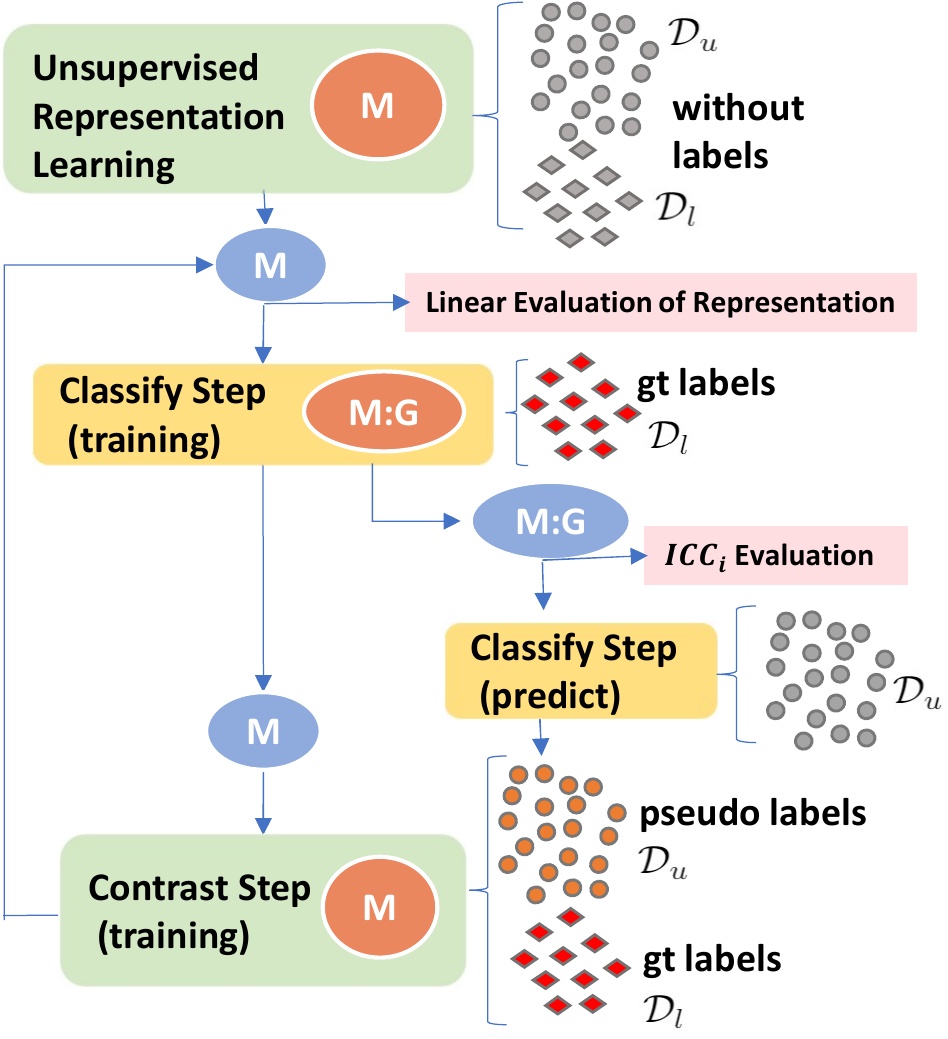}
\caption{Depiction of \textit{Iterative-Contrast-Classify} algorithm}
\label{fig:ICC-depiction}
\end{figure}

\section{Semi-Supervised Temporal Segmentation}\label{sec:semi-supervised-temporal-segmentation}
After unsupervised representation learning, the model $\mathbf{M}$ cannot yet be applied for action segmentation.  The decoder output must be coupled with a linear projection $\mathbf{G}$ and a soft-max to generate the actual segmentation.  $\mathbf{G}$ can only be learned using labels, \ie from $\mathcal{D}_L$, though the labels can be further leveraged to fine-tune $\mathbf{M}$ (Sec.~\ref{subsec:classify-step}).  Afterwards, $\mathbf{M}$ and $\mathbf{G}$ can be applied to unlabelled data $\mathcal{D}_U$ to generate pseudo-labels.  The pooled set of labels from $\mathcal{D}_L \cup \mathcal{D}_U$ can then be applied to update the representations in $\mathbf{M}$ (Sec.~\ref{subsec:contrast-step}).  By cycling between these updates, we propose an \emph{Iterative-Contrast-Classify (ICC)} algorithm (Sec.~\ref{subsec:Iterative-Contrast-Classify}) that performs semi-supervised action segmentation (see overview in Fig.~\ref{fig:ICC-depiction}).

\begin{table*}
\label{tab:unsupervised_linear}
\begin{center}
\small{
\begin{tabular}{l|ccccc|ccccc|ccccc}
    \hline
    & \multicolumn{5}{c|}{Breakfast} & \multicolumn{5}{c|}{50Salads} &  \multicolumn{5}{c}{GTEA}\\\hline
     & \multicolumn{3}{c}{$F1@\{10,25,50\}$} & Edit & MF & \multicolumn{3}{c}{$F1@\{10,25,50\}$} & Edit & MF &
     \multicolumn{3}{c}{$F1@\{10,25,50\}$} & Edit & MF \\
    \hline
    \rowcolor{LightGreen} Input I3D Baseline & 4.9 & 2.5 & 0.9 & 5.3 & 30.2 & 12.2 & 7.9 & 4.0 & 8.4 & 55.0 & 48.5 & 42.2 & 26.4 & 40.2 & 61.9 \\
    \rowcolor{LightGreen} Our Representations &\textbf{57.0} & \textbf{51.7} & \textbf{39.1} & \textbf{51.3} & \textbf{70.5}
     & \textbf{40.8} & \textbf{36.2} & \textbf{28.1} & \textbf{32.4} & \textbf{62.5} 
     & \textbf{70.8} & \textbf{65.0} & \textbf{48.0} & \textbf{65.7} & \textbf{69.1} \\
    \hline
    Improvement & 52.1 & 49.2 & 38.2 & 46.0 & 40.3 
                & 28.6 & 28.3 & 24.1 & 24.0 & 7.5 
                & 22.3 & 22.8 & 21.6 & 25.5 & 7.2 \\
    \hline
    \multicolumn{16}{c}{Our unsupervised learning gives a large improvement in segmentation compared to input features. 
    }\\
    \hline
    \rowcolor{LightCyan} Cluster & 11.7 & 8.0 & 3.9 & 12.2 & 36.1 
                      & 18.5 & 13.7 & 8.5 & 13.6 & 50.8
                      & 57.3 & 48.6 & 31.6 & 52.4 & 60.5\\
    \rowcolor{LightCyan} \textbf{(+)} Proximity & 24.4 & 19.2 & 11.5 & 21.3 & 50.0
                                & 18.6 & 13.5 & 8.0 & 13.5 & 51.6 
                                & 62.9 & 56.6 & 38.0 & 52.6 & 62.2\\
    \rowcolor{LightCyan} \textbf{(+)} Video-Level & 42.9 & 37.6 & 26.6 & 36.4 & 66.1  
                              & -- & -- & -- & -- & -- 
                              & -- & -- & -- & -- & -- \\
    \hline
    \multicolumn{16}{c}{Contribution of clustering and time-proximity conditions and video-level constraints for contrastive learning (with $\bz_6$
    ).}\\
    \hline
    \rowcolor{LightOrange} Last-Layer($\bz_6$) &  42.9 & 37.6 & 26.6 & 36.4 & 66.1 
                              & 18.6 & 13.5 & 8.0 & 13.5 & 51.6
                              & 62.9 & 56.6 & 38.0 & 52.6 & 62.2\\
    \rowcolor{LightOrange} Multi-Resolution($\feat$) & \textbf{57.0} & \textbf{51.7} & \textbf{39.1} & \textbf{51.3} & \textbf{70.5}
     & \textbf{40.8} & \textbf{36.2} & \textbf{28.1} & \textbf{32.4} & \textbf{62.5} 
     & \textbf{70.8} & \textbf{65.0} & \textbf{48.0} & \textbf{65.7} & \textbf{69.1} \\
    
    \hline
    Improvement & 14.1 & 14.1 & 12.5 & 14.9 & 4.4 
                & 22.2 & 22.7 & 20.1 & 18.9 & 10.9 
                & 7.9 & 8.4 & 10.0 & 13.1 & 6.9 \\
    \hline
    \multicolumn{16}{c}{Using Multi-Resolution($\feat$) representation instead of final decoder $\bz_6$ significantly improves learned representation scores.}\\
    \hline
    \end{tabular}
    }\end{center}
    \caption{Component-wise analysis of the unsupervised representation learning framework with a linear classifier.}
    \label{tab:unsupervised_ablation}
\end{table*}

\subsection{\textit{Classify Step: Learning $\mathbf{G}, \mathbf{M}$ with $\mathcal{D}_L$}}\label{subsec:classify-step}
Similar to the supervised C2F-TCN, each decoder layer's representations ${\bz}_u$ (temporal dim $\ceil*{T/2^{6-u}}$) is projected with a linear layer $\bG_u$ to $A$-dimensional vector where $A$ is the number of action classes. This is followed by a softmax to obtain class probabilities $\bp_{u}$ and a linear interpolation in time to up-sample back to the input length $T$. For frame $t$, the prediction ${\bp}[t]$ is a weighted ensemble of up-sampled $\bp_u$, \ie ${\bp}[t]\!=\!\sum_{u} \alpha_u \cdot \text{up}(\bp_u, T)$  where $\alpha_u$ is the ensemble weight of decoder $u$ with $\sum \alpha_u\!=\!1$, and $\text{up}\BK{\bp_u, T}$ denotes the upsampled decoder output of length $T$. The sum is action-wise and the final predicted action label is 
$\hat{y}_t = {\argmax}_{k \in \calA} \, {\bp}[t, k]$. Note that $\bG := \{\bG_u\}$ differs from the linear classifier $\bG_f$ of Sec.~\ref{subsec:linear-evaluation}.  $\bG_f$ is applied to the representation $\mathbf{f}$ used for the contrastive learning (see Sec.~\ref{subsec:multi-resolution-similarity}), whereas $\{\bG_u\}$ are applied individually to different $\bz_u$.

In addition to learning $\bG$, $\calD_L$ can be leveraged to fine-tune $\mathbf{M}$ as well.
In Eq.~\eqref{eqn:positive-negative}, the positive and negative sets $\calP_{n,i}$ and $\calN_{n,i}$ can be modified for $\calD_L$ to use the ground truth labels by replacing the unsupervised cluster labels $l_n[t^n_i]$ with ground truth action labels $y_n[t^n_i]$. Note that the learning rate used for fine-tuning the parameters of the model $\mathbf{M}$ is significantly lower than linear projection layers $\bG_u$. The loss used is $\calL = \losscross(\calD_L) + \losscons'(\calD_L)$, where $\losscons'$ is as defined in Eq.~$\eqref{eqn:contrastice-loss}$ but with $l_n[t^n_i]$ replaced by $y_n[t^n_i]$. 

\subsection{\textit{Contrast Step: Update $\mathbf{M}$ with $\mathcal{D}_U \cup \mathcal{D}_L$}}\label{subsec:contrast-step}
After fine-tuning, we can use %$\mathbf{M}_{\text{ss}}$ to 
$\mathbf{M}$ and $\mathbf{G}$ to
predict frame-level action labels $\hat{y}_n$ for any unlabelled videos, \ie pseudo-labels for $\calD_U$. This affords the possibility update the representation in $\mathbf{M}$. To that end, we again modify $\calP_{n,i}, \calN_{n,i}$ in Eq.~$\eqref{eqn:positive-negative}$ by replacing the cluster labels $l_n[t^n_i]$ with the pseudo-labels $\hat{y}_n[t]$ and ground truth labels $y_n[t^n_i]$ for $\calD_U$ and $\calD_L$ respectively.  $\mathbf{M}$ is then updated by applying the contrastive loss $\calL' = \losscons'(\calD_U) + \losscons'(\calD_L)$ where $\losscons'$ is as defined in Eq.~$\eqref{eqn:contrastice-loss}$.

\subsection{\textit{Iterative-Contrast-Classify} (ICC)}\label{subsec:Iterative-Contrast-Classify}
The pseudo labels for $\calD_U$ are significantly better representative of the (unseen) action labels than the clusters obtained from the input I3D features used in the unsupervised stage. Thus we can improve our contrastive representation by using the pseduo labels (obtained after \textit{classify}) for another \textit{contrast} step. This refined representation in turn can help in finding better pseudo labels through another following \textit{classify} step. By iterating between the contrast and classify in Secs.~\ref{subsec:classify-step} and~\ref{subsec:contrast-step} (see Fig.~\ref{fig:ICC-depiction}), we can thus progressively improve the performance of the semi-supervised segmentation. The segmentation performance is evaluated at the end of the \textit{classify} step after the training of $\mathbf{G}$.  We denote the combined model of $\mathbf{M}$ and $\mathbf{G}$ for each iteration $i$ as $\text{ICC}_{i}$.  In this way, initial unsupervised representation learning can be considered the \emph{``contrast''} step of $\text{ICC}_1$, where cluster labels are used instead of pseudo-labels.  We found that performance saturates after $4$ iterations of \textit{contrast-classify} and refer to the performance of $\text{ICC}_{4}$ as our final semi-supervised result.

\begin{table*}[t]
\centering
\small
\begin{tabular}{c|c|ccccc|ccccc|ccccc}
\hline
& & \multicolumn{5}{c|}{Breakfast} & \multicolumn{5}{c|}{50Salads} & \multicolumn{5}{c}{GTEA} \\
\hline
\%$D_L$ &\textbf{Method} & \multicolumn{3}{c}{$F1@\{10,25,50\}$} & Edit & MoF & \multicolumn{3}{c}{$F1@\{10,25,50\}$} & Edit & MoF & \multicolumn{3}{c}{$F1@\{10,25,50\}$} & Edit & MoF \\
\hline
\rowcolor{LightOrange} & $\text{ICC}_1$  
                    & 54.5 & 48.7 & 33.3 & 54.6 & 64.2 
                    & 41.3 & 37.2 & 27.8 & 35.4 & 57.3
                    & 70.3 & 66.5 & 49.5 & 64.7 & 66.0 \\
\rowcolor{LightOrange}
                    & $\text{ICC}_2$  
                    & 56.9 & 51.9 & 34.8 & 56.5 & 65.4 
                    & 45.7 & 40.9 & 30.7 & 40.9 & 59.5
                    & 77.0 & 70.6 & 54.1 & 67.8 & 68.0\\
\rowcolor{LightOrange} 
                    & $\text{ICC}_3$ 
                    & 59.9 & 53.3 & 35.5 & 56.3 & 64.2 
                    & 50.1 & 46.7 & 35.3 & 43.7 & 60.9
                    & 77.6 & 71.2 & 54.2 & 71.3 & 68.0\\
\rowcolor{LightOrange} 
\multirow{-4}{*}{\textbf{$\approx$5}} & $\text{ICC}_4$ 
                    & 60.2 & 53.5 & 35.6 & 56.6 & 65.3
                    & 52.9 & 49.0 & 36.6 & 45.6 & 61.3 
                    & 77.9 & 71.6 & 54.6 & 71.4 & 68.2 \\
                    
\cline{2-17}
& Gain 
                    & 5.7 & 4.8 & 2.3 & 2.0 & 1.1
                    & 11.6 & 11.8 & 8.8 & 10.2 & 4.0
                    & 7.6 & 5.1 & 5.1 & 6.7 & 2.2 \\
\hline
\multicolumn{17}{c}{{Progressive semi-supervised improvement with more iterations of ICC.}} \\
\hline
\rowcolor{LightGreen}
 & Supervised  
                    & 15.7 & 11.8 & 5.9 & 19.8 & 26.0 
                    & 30.5 & 25.4 & 17.3 & 26.3 & 43.1
                    & 64.9 & 57.5 & 40.8 & 59.2 & 59.7\\
\rowcolor{LightGreen} & Semi-Super 
                    & \textbf{60.2} & \textbf{53.5} & \textbf{35.6} & \textbf{56.6} & \textbf{65.3}
                    & \textbf{52.9} & \textbf{49.0} & \textbf{36.6} & \textbf{45.6} & \textbf{61.3} 
                    & \textbf{77.9} & \textbf{71.6} & \textbf{54.6} & \textbf{71.4} & \textbf{68.2} \\
\cline{2-17}
\multirow{-3}{*}{\textbf{$\approx$5}} & Gain 
                    & 44.5 & 41.7 & 29.7 & 36.8 & 39.3 
                    & 22.4 & 23.6 & 19.3 & 19.3 & 18.2
                    & 13.0 & 14.1 & 13.8 & 12.2 & 8.5
                    \\
\hline
 \rowcolor{LightGreen} &  Supervised 
        & 35.1 & 30.6 & 19.5 & 36.3 & 40.3 
        & 45.1 & 38.3 & 26.4 & 38.2 & 54.8 
        & 66.2 & 61.7 & 45.2 & 62.5 & 60.6 \\ 
 \rowcolor{LightGreen} & Semi-Super 
        & \textbf{64.6} & \textbf{59.0} & \textbf{42.2} & \textbf{61.9} & \textbf{68.8} 
        & \textbf{67.3} & \textbf{64.9} & \textbf{49.2} & \textbf{56.9} & \textbf{68.6} 
        & \textbf{83.7} & \textbf{81.9} & \textbf{66.6} & \textbf{76.4} & \textbf{73.3} \\
\cline{2-17}
\multirow{-3}{*}{\textbf{$\approx$10}} & Gain
        & 29.5 & 28.4 & 22.7 & 25.6 & 28.5 
        & 22.2 & 26.6 & 22.8 & 18.7 & 13.8 
        & 17.5 & 20.2 & 21.4 & 13.9 & 12.7\\
\hline
\multicolumn{17}{c}{{Semi-Super (our $\text{ICC}_4$) significantly improves supervised counterpart using same labelled data amount.  See also Fig.~\ref{fig:teaser}}}\\
\hline
\rowcolor{LightCyan} & Supervised*  
                    & 69.4 & 65.9 & 55.1 & 66.5 & 73.4 
                    & 75.8 & 73.1 & 62.3 & 68.8 & 79.4 
                    & 90.1 & 87.8 & 74.9 & 86.7 & 79.5 \\

 \rowcolor{LightCyan} & Semi-Super
             & \textbf{72.4} & \textbf{68.5} & \textbf{55.9} & \textbf{68.6} & \textbf{75.2} 
             & \textbf{83.8} & \textbf{82.0} & \textbf{74.3} & \textbf{76.1} & \textbf{85.0} 
             & \textbf{91.4} & \textbf{89.1} & \textbf{80.5} & \textbf{87.8} & \textbf{82.0} \\
\cline{2-17}
\multirow{-3}{*}{\textbf{100}} 
            & Gain & 3.0 & 2.6 & 0.8 & 2.1 & 1.8 
            & 8.0 & 8.9 & 12.0 & 7.3 & 5.6 
            & 1.3 & 1.3 & 5.6 & 1.1 & 2.5 \\
\hline
\multicolumn{17}{c}{{Semi-Super (our $\text{ICC}_4$) helps improve fully supervised learning. * reported from C2F-TCN without test-augment-action-loss.}} \\
\hline
\end{tabular}
\caption{{Our final all metrics evaluation of proposed ICC algorithm on 3 benchmark action segmentation datasets.}}\label{tab:icc_improvement}
\end{table*}

\section{Experiments}\label{sec:experiments}

\subsection{Datasets, Evaluation, Implementation Details}\label{subsec:eval-details} 

We test on Breakfast Actions~\cite{kuehne2014language} (1.7k videos, 10 complex activities, 48 actions), 50Salads~\cite{stein2013combining} (50 videos, 19 actions) and GTEA~\cite{GTEA-fathi2011learning} (28 videos, 11 actions). The standard evaluation criteria are the Mean-over-Frames (MoF), segment-wise edit distance (Edit), and $F1$-scores with IoU thresholds of $0.10$, $0.25$ and $0.50$ ($F1@\{10, 25, 50\}$). 
We report results with the stronger I3D input features and make comparisons with IDT features in Appendix-D.

We use the specified train-test splits for each dataset and randomly select $5\%$ or $10\%$ of videos from the training split for labelled dataset $\calD_L$. As GTEA and 50Salads are small, we use $3$ and $5$ videos as 5\% and 10\% respectively to
incorporate all $A$ actions. 
We report mean and standard deviation of five different selections. For unsupervised representation learning, we use all the unlabelled videos in the dataset which is in line with other unsupervised works~\cite{unsupervised-kukleva2019unsupervised, selfsupervised-chen2020action}.
We sample frames from each video with $K=\{20, 60, 20\}$ partitions,
{$\epsilon\!\!\approx\!\!\frac{1}{3K}$} for sampling, and temporal proximity $\delta\!=\!\{0.03, 0.5, 0.5\}$ for Breakfast, 50Salads, and GTEA respectively. 
The contrastive temperature $\tau$ in Eqs.~\eqref{eqn:frame-level-contrast} and \eqref{eqn:video-level-contrast} is set to $0.1$. We also leverage the feature augmentations of C2F-TCN, with details and ablations in Appendix-B.1. 

\subsection{Evaluation of Representation Learning}\label{subsec:eval-learned-repr}
\subsubsection{Linear Classification Accuracy} of our unsupervised representation learning (see Sec.~\ref{subsec:linear-evaluation}) is shown in Table~\ref{tab:unsupervised_ablation}.  In the first green section, we evaluate the input I3D features with a linear classifier to serve as a baseline. Our representation has significant gains over the input I3D, verifying the ability of the base TCN to perform the task of segmentation with our designed unsupervised learning.

\subsubsection{Frame- and Video-Level Contrastive Learning:}
The blue section of Table~\ref{tab:unsupervised_ablation} breaks down the contributions from Sec.~\ref{sec:frame_level_contratsive} and \ref{subsec:video-level-contrast} when forming the positive and negative sets of contrastive learning from Eq.~$\eqref{eqn:positive-negative}$. The `\textit{Cluster}' row applies cluster labels condition \ie $l_n[t^n_i] = l_m[t^m_j]$ and `\textit{ (+) Proximity}' adds the condition $|t^n_i - t^m_j| < \delta$.
Adding time proximity is more effective for Breakfast and GTEA likely because their videos follows a more rigid sequencing than 50Salads. Adding the \textit{Video-Level} contrastive loss from Sec.~\ref{subsec:video-level-contrast} in Breakfast gives further boosts. 

\subsubsection{Multi-Resolution Representation:} The orange section of Table~\ref{tab:unsupervised_ablation} verifies that our multi-resolution representation $\feat$ (see Sec.~\ref{subsec:multi-resolution-similarity}) outperforms the use of only the final decoder layer feature $\bz_6$.  Gains are especially notable for the F1-score and Edit-distance, verifying that $\mathbf{f}$ has less over-segmentation. 

\subsubsection{No Initial Representation Learning:} Although model $\mathbf{M}$ is continually updated during the ICC, Appendix-D.2 shows that the initial unsupervised representation learning is critical. Bypassing this step in the first iteration of ICC and going straight to \emph{classify} step results in a gap of $\ge$ 10\% F1.

\subsection{Evaluation of Semi-Supervised Learning}

\subsubsection{ICC Components:} The orange section of Table~\ref{tab:icc_improvement} shows the progressive improvements as we increase the number of iterations of our proposed ICC algorithm. The gain in performance is especially noticeable for Edit and F1 scores. 
Segmentation result reported are after the \textit{classify} step.  Improvements gained by updating the feature representation after the \textit{contrast} step but before \textit{classify} of the next iteration is shown in the Appendix-D.3.

\subsubsection{Additional Ablations:} Due to lack of space, we refer the reader to Appendix-B. Notably, we demonstrate ICC results with alternative base model ED-TCN~\cite{TED-lea2017temporal} and draw difference to using MS-TCN~\cite{li2020ms}.

\subsubsection{Semi-Supervised vs Supervised:}
The green and blue sections of Table~\ref{tab:icc_improvement} shows 
our final \emph{`Semi-Super'} results, \ie $\text{ICC}_4$ for various percentages of labelled data.  We compare with the \textit{`Supervised'} case of training the base model C2F-TCN with the same labelled dataset $\calD_L$; for fairness, 100\% C2F-TCN results is reported without test-time augmentations and action loss. ICC significantly outperforms the supervised baseline in all the metrics (see also Fig.~\ref{fig:teaser}) for all amounts of training data, including 100\%. 

In fact, with just 5\% of labelled videos, we are only 8\% less in MoF in Breakfast actions compared to fully-supervised (100\%). Using less than 5\% (3 videos for 50salads and GTEA) of training videos makes it difficult to ensure coverage of all the actions.

\subsubsection{Comparison to \textit{SOTA} and differences:} Our work being the first to do semi-supervised temporal action segmentation, is not directly comparable to other works. Table~\ref{tab:different_supervsion_breakfast} shows our MoF is competitive with other forms of supervision on all three datasets. 
TSS and SSTDA uses weak labels for \textit{all} videos vs. our work requiring full labels for only few videos.  

\begin{table}

\begin{center}
\small{
    \begin{tabular}{c|c|ccc}
    \hline
     & Method & Breakfast & 50salads & GTEA \\\hline
     \rowcolor{LightCyan} & MSTCN'20 & 67.6 & 83.7 & 78.9 \\
    \rowcolor{LightCyan} & SSTDA'20  & 70.2 & 83.2 & 79.8 \\
    \rowcolor{LightCyan} & *C2F-TCN'21  & 73.4 & 79.4 & 79.5 \\
    \cline{2-5}
    \rowcolor{LightCyan} \multirow{-4}{*}{\textbf{Full}} & \textbf{Ours ICC (100\%)} & \textbf{75.2} & \textbf{85.0} & \textbf{82.0} \\
    \hline
    \rowcolor{LightPink} & SSTDA(65\%)  & 65.8 & 80.7 & 75.7\\
     \rowcolor{LightPink} \multirow{-2}{*}{\textbf{Weakly}} & TSS'21 & 64.1 & 75.6 & 66.4 \\
    \hline
    \rowcolor{LightGreen} & {Ours ICC (40\%)} & 71.1 & 78.0 & 78.4 \\
    \rowcolor{LightGreen} & {Ours ICC (10\%)} & 68.8 & 68.6 & 73.3 \\
     \rowcolor{LightGreen} \multirow{-3}{*}{\textbf{Semi}} & {Ours ICC (5\%)}  & 65.3 & 61.3 & 68.2 \\\hline
    \end{tabular}
}\end{center}
\caption{Segmentation MoF comparison with \textit{SOTA} on 3 benchmark datasets. Our ICC can improve its fully-supervised counterpart. Our semi-supervised results is competitive in MoF with different levels of supervision.}
\label{tab:different_supervsion_breakfast}
\end{table}

\section{Conclusions} 

In our work we show that pre-trained input features that capture semantics and motion of short-trimmed video segments can be used to learn higher-level representations to interpret long video sequences. Our proposed multi-resolution representation formed with outputs from multiple decoder layers, implicitly bring temporal continuity and consequently large improvements in unsupervised contrastive representation learning. Our final semi-supervised learning algorithm ICC can significantly reduce the annotation efforts, with 40\% labelled videos approximately achieving fully-supervised (100\%) performance. Furthermore, ICC also improves performance even when used with 100\% labels.

\appendix
\section{Appendix}

In section~\ref{sec:implementation_details} we details our training hyper-parameters used for unsupervised and semi-supervised setup. We also provide variations in results with change in important hyper-parameters of our algorithm to experimentally validate our choice of hyper-parameters. In section~\ref{sec:base_architecture} we provide additional detail of the base architecture C2F-TCN~\cite{singhania2021coarse} used and validate the choice with respect to other base model like ED-TCN~\cite{TED-lea2017temporal}, MSTCN~\cite{li2020ms}. In section~\ref{sec:multi_resolution_feature} we experimentally validate the normalization strategy used for our multi-resolution feature and show derivation of temporal continuity implicitly incorporated in it. In section~\ref{sec:additional_ablation} we show additional experimental and qualitative visualization results. We show visualization of representation learnt and input I3D feature, and show results with IDT as input features in section~\ref{subsec:IDT}. The importance of our unsupervised representation learning step in final results is shown in section~\ref{subsec:icc_wo_pretraining}. We show quantitative and qualitative improvements in representations learnt and semi-supervised classification results with multiple iteration of ICC in section ~\ref{subsec:icc_progression_results}. Finally we report deviations of our results with different selections of labelled videos in ~\ref{subsec:mean_std_results}.

\section{Implementation Details \& Hyperparameter Selection}\label{sec:implementation_details}

All our experiments are conducted on \textit{Nvidia GeForce RTX 2080 Ti} GPUs with 10.76 Gb memory. We use the Adam optimizer with Learning Rate (LR), Weight Decay (WD), Epochs (Eps), and Batch Size (BS) used for our unsupervised and semi-supervised setup shown in Table \ref{tab:hyperparameters}. For the \textit{contrast step} of $\text{ICC}_i$ for $i \ge 2$, we reduce the learning rate to be 0.1 times that of the \textit{contrast step} learning rate used in the first iteration (\ie $\text{ICC}_1$). We use a higher number of epochs for the semi-supervised fine-tuning step, but as the number of labeled data is minimal, the training time is quite less. For all three datasets, we follow the $k$-fold cross-validation averaging to report our final results of representation, semi-supervised and fully-supervised results(however, we tune our model hyper-parameters for a hold-out validation set from training dataset). Here $k=\{4, 5, 4\}$ for Breakfast, 50Salads and GTEA respectively and is same as used in other works ~\cite{li2020ms, selfsupervised-chen2020action}.

\subsection{Sampling strategy, number of samples $2K$} We show ablations for the choice of $2K$ \ie number of representation samples drawn per video as described in section 4.1 of main paper in Table~\ref{tab:num_samples}. Thus our value of $2K$ is determined by experimental validation. In 50salads with $2K=120$ and batch size of 50, we get roughly 0.6 million positive samples per batch, with each positive sample having roughly around 6.5K negative samples. 

\begin{table}[t]
    \begin{center}
        \small{
    
    \begin{tabular}{l|ccccc}
    \hline
         Samples($2K$) & \multicolumn{3}{c}{$F1\{10, 25,50\}$} & Edit & MoF \\\hline
         60 & 38.5 & 33.2 & 25.1 & 29.9 & 62.5\\
         120 & 40.8	& 36.2 & 28.1 & 32.4 & 62.5\\
         180 & 39.1	& 34.5 & 27.3 & 30.8 & 62.1\\\hline
    \end{tabular}
    }\end{center}
    \caption{Ablation results of \textbf{number of samples per video} required for representation learning.}
    \label{tab:num_samples}
\end{table}

\begin{table}[]
    \begin{center}
    \small{
    \begin{tabular}{l|l|ccccc}
    \hline
     \%$\calD_L$ & \textbf{Method} & \multicolumn{3}{c}{$F1@\{10,25,50\}$} & Edit & MoF\\\hline
        100\% & Full-Supervised & 68.0 & 63.9 & 52.6 & 52.6 & 64.7 \\
        \hline
         5\% & Supervised &  32.4 & 26.5 & 14.8 & 25.5 & 39.1 \\
         5\% & our ICC &  39.3 & 34.4 & 21.6 & 32.7 & 46.4 \\
         \cline{2-7}
         5\% & Gain & 6.9 & 7.9 & 6.8 & 7.2 & 7.3 \\
         \hline
    \end{tabular}
    }\end{center}
    \caption{\textbf{Our semi-supervised (final $\text{ICC}_4$) results with EDTCN} \cite{TED-lea2017temporal} on 50salads with 5\% labelled data significantly improves over its supervised counterpart.}\label{tab:EDTCN_semi_super}
\end{table}

\begin{table*}[]
\begin{center}
    \small{
    % \begin{tabular}{p{2.4cm}|p{0.6cm}p{0.6cm}p{0.4cm}p{0.3cm}| p{0.6cm}p{0.6cm}p{0.4cm}p{0.3cm}|p{0.6cm}p{0.6cm}p{0.4cm}p{0.3cm}}
    \begin{tabular}{c|cccc|cccc|cccc}
    \hline
        & \multicolumn{4}{c|}{Breakfast} 
        & \multicolumn{4}{c|}{50Salads} 
        & \multicolumn{4}{c}{GTEA}\\
        \cline{2-13}
        Algorithm & LR & WD & Eps. & BS 
                  & LR & WD & Eps. & BS
                  & LR & WD & Eps. & BS\\
        \hline
        \textit{Contrast step} (model $\mathbf{M}$) & 1e-3 & 3e-3 & 100 & 100 
                     & 1e-3 & 1e-3 & 100 & 50 
                     & 1e-3 & 3e-4 & 100 & 30 \\
        \hline
        \textit{Classify step} (classifier $\bG$) & 1e-2 & 3e-3 & 700 & 100
                            & 1e-2 & 1e-3 & 1800 & 5 
                            & 1e-2 & 3e-4 & 1800 & 5 \\
        \textit{Classify step} (model $\mathbf{M}$) & 1e-5 & 3e-3 & 700 & 100
                                  & 1e-5 & 1e-3 & 1800 & 5 
                                  & 1e-5 & 3e-4 & 1800 & 5 \\
    \hline
    \end{tabular}
    }\end{center}
    \caption{Training hyperparameters learning rate(LR), weight-decay(WD), epochs(Eps.) and Batch Size(BS) used for different datasets for unsupervised and semi-supervised learning.}\label{tab:hyperparameters}
\end{table*}

\begin{table}[]
    \begin{center}
    \small{
    \begin{tabular}{cccc}
        \hline
         Cluster-Type (Num-Clusters) & Breakfast & 50Salads & GTEA \\
         \hline
         FINCH ($A$)  & 61.7 & 56.3 & 56.7 \\
         Kmeans ($A$) & 70.0 & 60.4 & 65.6 \\
         \textbf{Kmeans ($\approx 2A$)} & \textbf{70.5} & \textbf{62.5} & \textbf{69.1} \\
        \hline
    \end{tabular}
    }\end{center}
    \caption{Unsupervised Representation's MoF variation with different clustering types and number of clusters used during training. $A$ denotes number of unique actions in the dataset.}
    \label{tab:kmeans_finch}
\end{table}

\subsection{Input feature clustering} 
In section 4.1 of the main paper, unsupervised feature learning requires cluster labels from the input features. 
We cluster at the mini-batch level with a standard {$k$-means} and then compare with Finch~\cite{sarfraz2019efficient}, an agglomerative clustering, that have shown to be useful in the unsupervised temporal segmentation \cite{sarfraz2021temporally}. 
Comparing the two in Table~\ref{tab:kmeans_finch}, we observe that $K$-means performs better.  We speculate that this is because Finch is designed for per-video clustering. In contrast, our clustering on the mini-batch is on a dataset level, i.e., over multiple video sequences of different complex activities. 

For choosing $k$ in the $k$-Means clustering, we choose $\approx 2A$ ($A$ denotes number of unique actions) number of clusters resulting in $K=\{100, 40, 30\}$ for Breakfast, 50Salads, and GTEA datasets, respectively.  The advantage of using  $\approx 2A$ clusters versus simply $A$ is verified in Table~\ref{tab:kmeans_finch}. The improvement is greater for datasets with fewer action classes like GTEA and 50salads than the Breakfast action dataset.

\section{Details And Choice Of Base Architecture}\label{sec:base_architecture}
\subsection{Details of C2F-TCN}
Our base architecture is same as C2F-TCN~\cite{singhania2021coarse}, which takes snippet-level features like IDT or I3D as input and which has convolution block used for encoder-decoder of kernel size $3$ with total of $4.07$ million trainable parameters. To train the C2F-TCN, we use %\textit
{downsampled} input features and a %\textit
{multi-resolution augmentation} training strategy as recommended by \cite{singhania2021coarse}. We briefly discuss downsampling and augmentation used here and also show its effect on our unsupervised feature learning in Table \ref{tab:feature-agument}.

\textbf{Downsampling:}
The pre-trained input feature representations i.e I3D and IDT referenced as $\vid \in \bbR^{T \times F}$ and the ground truth $y$ of temporal dimension $T$ is downsampled by $w$ to obtain the input feature vector $\vid_{in} \in \bbR^{T_{in} \times F}$ and ground truth $y_{in}$ of temporal dimension $T_{in}$, where $T_{in}= \frac{T}{w} $. This is done by max-pooling the features using a temporal window of size $w$ and assigning the corresponding label which is most frequent in the given window.  Specifically, if the input feature and ground truth label at time $t$ i.e $\vid^{w}_{in}[t]$ and $y^{w}_{in}[t]$ for some temporal window $w > 0$ is, then the downsampled feature and label can be defined as 
\begin{align}\label{eqn:downsample}
    \vid^{w}_{in}[t] &= \max_{\tau \in \left[wt, wt+w\right)} \vid[\tau] \\ 
    y^{w}_{in}[t] &= \underset{k \in \calA}{\argmax} \sum_{\tau=wt}^{wt+w} \bbI\sqBK{y\left[\tau\right]=k}.
\end{align}

\textbf{Temporal feature augmentation} applies max-pooling 
over multiple temporal resolution of the features by varying the window size $w$ uniformly from $[w_{min}, w_{max}]$ during training.

We use the same values of $w_0$ as~\cite{singhania2021coarse} with $w_0\!=\!\{10, 20, 4\}$ for Breakfast, 50Salads and GTEA respectively and $w_{min}\!=\!\frac{1}{2} w_0$   and $w_{max}\!=\!2 \times w_0$. In Table~\ref{tab:feature-agument} we show the improvements in unsupervised features linear evaluation scores when training with feature-augmentation. We do not use test time feature augmentations and all results in our paper is reported without test time augmentations.

\begin{table*}[t]

\begin{center}
\small{
\begin{tabular}{l|ccccc|ccccc|ccccc}
\hline
& \multicolumn{5}{c|}{Breakfast} 
& \multicolumn{5}{c|}{50Salads} 
& \multicolumn{5}{c}{GTEA}\\
\hline
\textbf{Method} &
\multicolumn{3}{c}{$F1\{10, 25,50\}$} & Edit & MoF & \multicolumn{3}{c}{$F1\{10, 25,50\}$} & Edit & MoF &
\multicolumn{3}{c}{$F1\{10, 25,50\}$} & Edit & MoF \\
\hline
No Augment & 55.6 & 50.2 & 36.5 & 49.4 & 69.4 & 
              40.0 & 34.1 & 27.0 & 31.0 & 62.3 &
              70.0 & 63.4 & 47.2 & 65.6	& 69.0 \\
Augment & \textbf{57.0} & \textbf{51.7} & \textbf{39.1} & \textbf{51.3} & \textbf{70.5} & 
 \textbf{40.8} & \textbf{36.2} & \textbf{28.1} & \textbf{32.4} & \textbf{62.5} &
 \textbf{70.8} & \textbf{65.0} & \textbf{48.0} & \textbf{65.7} & \textbf{69.1} \\

\hline
\end{tabular}
}\end{center}
\caption{\textbf{Impact of using Multi-Resolution Augmentation}} \label{tab:feature-agument}
\end{table*}

\begin{table*}[h]

 \begin{center}
    \small{
\begin{tabular}{l|ccccc|ccccc|ccccc}
\hline
& \multicolumn{5}{c|}{Breakfast} 
& \multicolumn{5}{c|}{50Salads} 
& \multicolumn{5}{c}{GTEA}\\
\hline
\textbf{Method} &
\multicolumn{3}{c}{$F1\{10, 25,50\}$} & Edit & MoF & \multicolumn{3}{c}{$F1\{10, 25,50\}$} & Edit & MoF &
\multicolumn{3}{c}{$F1\{10, 25,50\}$} & Edit & MoF \\
\hline

Alternate $\feat^{'}[t]$ &44.3 & 38.3 & 26.1 & 40.9 & 60.9 &
                          32.9 & 27.3 & 19.9 & 26.5 & 51.2 &
                          56.4 & 48.6 & 31.3 & 52.1 & 58.9\\
\textbf{Proposed} $\feat[t]$ & \textbf{57.0} & \textbf{51.7} & \textbf{39.1} & \textbf{51.3} & \textbf{70.5} & 
                     \textbf{40.8} & \textbf{36.2} & \textbf{28.1} & \textbf{32.4} & \textbf{62.5} &
                     \textbf{70.8} & \textbf{65.0} & \textbf{48.0} & \textbf{65.7} & \textbf{69.1} \\

\hline
\end{tabular}
}\end{center}
\caption{\textbf{Importance of normalization order for Multi-Resolution Feature}}\label{tab:multi-res-feature} \label{tab:feature_similarity_formation}
\end{table*}

\begin{algorithm}[t]
\caption{Iterative \textit{Contrast-Classify} algorithm}\label{algo:contrast-classify}
\begin{algorithmic}[1]
% \Comment{\emph{Contrast-classify}}
\While {\textit{iter} $\le$ MaxIter}
    \For {\emph{epoch} $\le$ MaxEpoch} \Comment{\textbf{\emph{Contrast}}:}
    \State sample minibatch $\{\vid_n\}_{n=1}^N \subset \calD_u \cup \calD_l$
        \If {\emph{iter} $= 0$}
            \State $\{l_n\}_{n=1}^N \gets $ \emph{Cluster}$\big(\{\vid_n\}_{n=1}^N\big)$
        \EndIf
        \State $\losscons \gets$ \emph{Contrastive}$\big(\{\BK{\vid_n, l_n}\}_{n=1}^N, \mathbf{M}\big)$
        \State minimize $\losscons$ and update $\mathbf{M}$
    \EndFor
    \For {$\vid_n \in \calD_l$}
        \State $l_n \gets y_n$
    \EndFor
    \For {\emph{epoch} $\le$ MaxEpoch}\Comment{\textbf{\emph{Classify}}:} 
        \State sample minibatch $\{\vid_n\}_{n=1}^N \subset \calD_l$
        \State $\losscons \gets$ \emph{Contrastive}$\big(\{\BK{\vid_n, l_n}\}_{n=1}^N, \mathbf{M}\big)$
        \State $\{\bar{\bp}_n\}_{n=1}^N \gets$ \emph{Predict}$(\{\vid_n\}_{n=1}^N, \bG)$
        \State $\losscross \gets$ \emph{CrossEntropy}$\big(\{\BK{\bar{\bp}_n, y_n}\}_{n=1}^N\big)$
        \State $\calL \gets \losscons + \losscross$
        \State minimize $\calL$ and update $\mathbf{M}$ and $\bG$
    \EndFor
    \For {$\vid_n \in \calD_u$}
        \State $l_n \gets$ \emph{Predict}$(\vid_n, \bG)$
    \EndFor
    \State \emph{iter} $\gets$ \emph{iter} $+ 1$
\EndWhile
\end{algorithmic}\label{algo:contrast_classify}
\end{algorithm}

\subsection{Other baseline models}

We try our entire algorithm~\ref{algo:contrast_classify} for semi-supervised temporal segmentation (outlined in section 5, Figure 4 of main paper) with unsupervised representation learning and iterative contrast classify with two other base TCN models: ED-TCN(an encoder-decoder architecture) and MSTCN (wavenet like refinement architecture).

\textbf{ED-TCN:} We show our proposed ICC to be working with ED-TCN~\cite{TED-lea2017temporal} in Table~\ref{tab:EDTCN_semi_super}, where ICC improves performance over its supervised counterpart. Due to smaller capacity of ED-TCN compared to C2F-TCN (indicated by the fully supervised performance (top row of table~\ref{tab:EDTCN_semi_super}) of ED-TCN 64.7\% vs C2F-TCN 79.4\% MoF), ICC algorithm improvement with ED-TCN is lower than improvement with C2F-TCN. For constrastive framework, an increased  performance with better model capacity has also been shown before in SimCLR~\cite{simCLR}.

\textbf{MSTCN:} Our proposed algorithm does not work well with MSTCN~\cite{li2020ms} architecture, possibly due to the fact that MSTCN is not designed for representation learning as 3 out of the 4 model blocks consists of refinement stages, where each stage takes \textit{class probability} vectors as input from the previous stages. Therefore, representation learning and classifier cannot be decoupled, making alternative classifier-representation learning algorithm impossible. Further, MS-TCN do not have multiple temporal resolution representation like encoder-decoder architecture which play a significant role in our contrastive learning as discussed earlier. 
\begin{table}[t]
    \begin{center}
    \small{
    \begin{tabular}{lcc}
    \hline
    \textbf{Representation} & \textbf{Breakfast} & \textbf{50Salads} \\
    \hline
    IDT + CTE-MLP & 45.0$^*$ & 34.1$^*$ \\\hline
    
    IDT & 18.2 & 36.9 \\
    
    \textbf{Ours} (Input IDT) & 67.9 & 52.2 \\
    \hline
    Gain  & +49.7 & +15.3 \\
    \hline
    \end{tabular}
    }\end{center}
    \caption{Our learned representation linear evaluation of segmentation's  MoF significantly improves upon \textbf{input IDT features}. $^*$ indicates evaluations based on publicly available checkpoints.}
    \label{tab:unsupervised_sota}
\end{table}

\section{Multi-Resolution Features} \label{sec:multi_resolution_feature}
\textbf{Normalization:}
Our proposed multi-resolution feature, as outlined in Section 4.3 of the main paper, is defined for frame $t$ as $\feat[t] = \BK{\bar{\bz}_1[t]:\bar{\bz}_2[t]:\ldots:\bar{\bz}_6[t]}$, where $\bar{\bz}_u[t] = {\hat{\bz}}_u[t]/\norm{{\hat{\bz}}_u[t]}$, \ie ${\hat{\bz}}_u[t]$, the upsampled feature from decoder $u$, is normalized first for each frame and then concatenated along the latent dimension.

\paragraph{Inherent temporal continuity of our feature $\feat$} The inherent temporal continuity encoded in our multi-resolution feature is discussed in Section 4.3 of the main paper.  For the `\textit{nearest}' neighbor upsampling strategy, the multi-resolution feature $\feat$ has the property of being similar for nearby frames. Coarser features like $\{\bz_1, \bz_2, \bz_3\}$ are more similar than fine-grained features at higher decoder layers. This also gives independence to the higher resolution features to have high variability even for nearby frames.
Specifically, for two frames $t, s \in \bbN$, if $\floor*{t/2^u} = \floor*{s/2^u}$ for some integer $u > 0$ then $\simi{\feat[t]}{\feat[s]} \ge 1 - u/3$. This follows from the fact that for nearest upsampling, $\floor*{t/2^u} = \floor*{s/2^u}$ for some $0 \le u \le 5$, implies that
\begin{align}\label{eq.nearest_meaning}
    \bz_v[t] = \bz_v[s] \quad\text{for all}\quad 1\le v \le 6-u.
\end{align}
Meaning, the lower resolution features coincide for proximal frames. As discussed and shown in equation (6) of the main text, all the layers have equal contribution while calculating similarity of our multi-resolution feature. For $t,s$, with $\floor*{t/2^u} = \floor*{s/2^u}$ for some $0 \le u \le 5$, we start with the equation (6) of main text to derive --
\begin{align*}
    & \simi{\feat[t]}{\feat[s]}\\ 
    &= \sum_{v=1}^6 \frac{1}{6} \cdot \simi{\bz_v[t]}{\bz_v[s]}\\
    &= \sum_{v=1}^{6-u} \frac{1}{6} \cdot \simi{\bz_v[t]}{\bz_v[s]} + \sum_{v=7-u}^{6} \frac{1}{6} \cdot \simi{\bz_v[t]}{\bz_v[s]}\\
    &= \frac{6-u}{6} + \sum_{v=7-u}^{6} \frac{1}{6} \cdot \simi{\bz_u[t]}{\bz_u[s]} \quad \BK{\text{from \eqref{eq.nearest_meaning}}}\\
    &\ge \frac{6-u}{6} - \frac{u}{6} \qquad\qquad \BK{\text{as}\,\, \simi{\cdot}{\cdot} \ge -1}\\
    &= 1 - \frac{u}{3}.
\end{align*}
That means for $\floor*{t/2^u} = \floor*{s/2^u}$ for some $0 \le u \le 5$ implies $\simi{\feat[t]}{\feat[s]} \ge 1 - \frac{u}{3}$. The inequality is trivial for $u > 5$.

\section{Additional Ablation}\label{sec:additional_ablation}

\subsection{Unsupervised Representation learning}\label{subsec:IDT}
\begin{figure*}[t]
    \centering
    \includegraphics[width=0.8\linewidth]{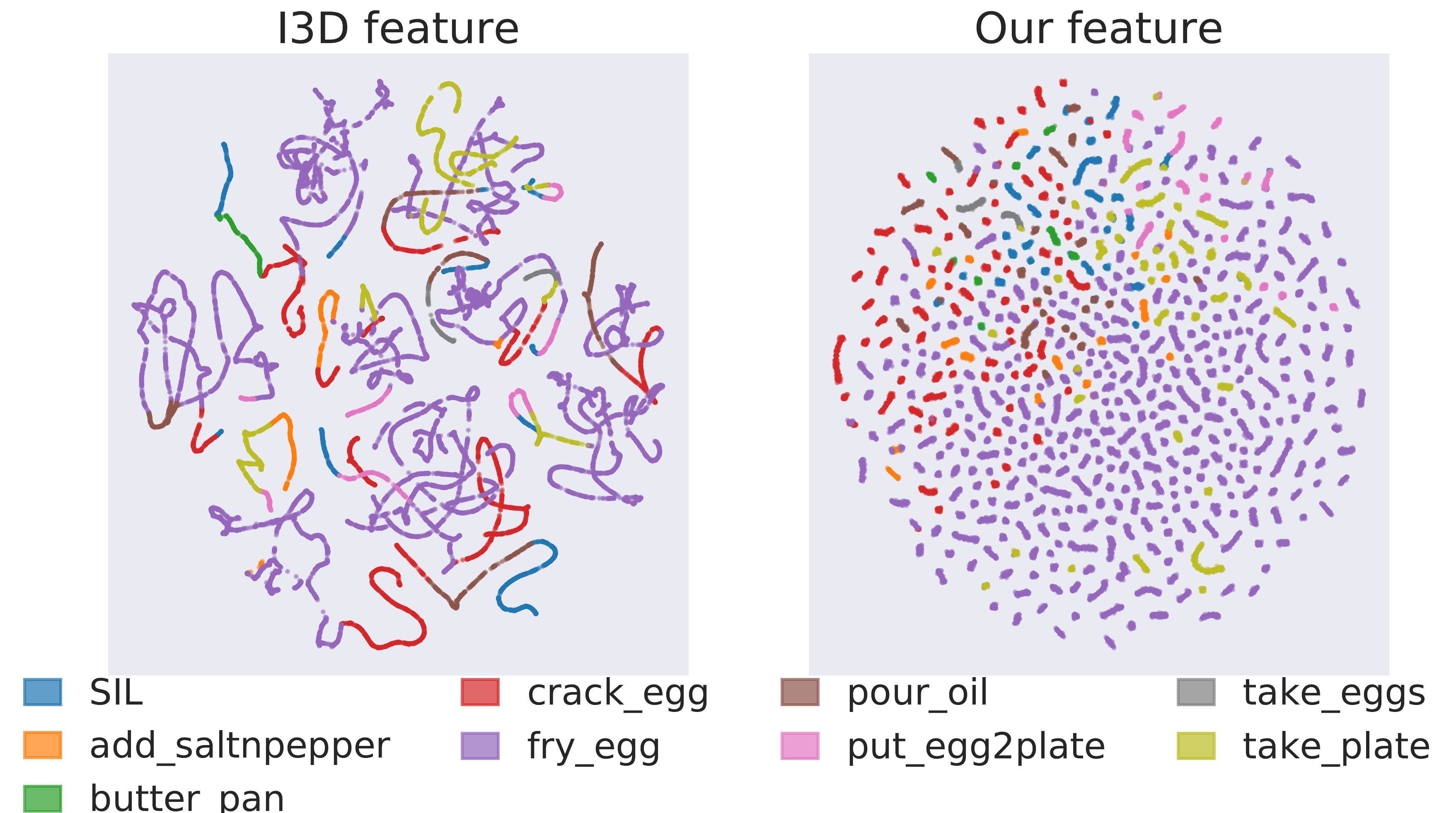}
    \caption{Change in UMAP scatter plot~\cite{UMAP} after our unsupervised representation learning. The plot is of seven videos of the complex activity ``\textit{fry\_egg}". Each point in the plot denotes a frame of a video while different colors represent different actions (specified in the legend). Left subplot shows the discriminativeness of input I3D features, while the right one shows representation obtained by our unsupervised learning. In the left figure, each of the connected path-like blobs belong to different videos. Meaning the I3D features of the frames of the same video are well-connected in the feature space and suggest that temporal continuity is the primary factor behind the distinction in the feature space. In contrast (in the right figure), our learned representation separates actions across different videos. There are also some locally-connected components in the right figure which actually belong to the frames of the same video which are of the same action as well. Hence, rather than separating frames primarily based on temporal continuity (like the I3D feature) our learned feature primarily separates based on action, and then locally on the basis of temporal continuity (same video).}
    \label{fig:Umap_representation}
\end{figure*}
\subsubsection{Visualization of representations learnt:} In Table 1 of the main paper we show the representations results for I3D features as input to C2F-TCN. In Figure~\ref{fig:Umap_representation} we visualize our learnt representation versus input I3D features with the help of UMAP(used for visualization by reducing data to 2-dimensions) visualization. We show our learnt representation incorporate separability based of actions and therefore have much higher segmentation linear classifier scores that input I3D features(Table 1 of main paper).

\subsubsection{Experiment Result with IDT features:} We omitted representations segmentation results when using IDT features as inputs to C2F-TCN instead of I3D due to space limitations and include it here in Table~\ref{tab:unsupervised_sota}. IDT features are obtained in unsupervised way with use of PCA and Gaussian Mixture Models~\cite{wang2013action} compared to I3D obtained from pre-trained models. We show that trends in improvement of our representation compared to input IDT features are similar to the results of Table 1 of the main paper.

\begin{table}[]
    \begin{center}
    \small{
    \begin{tabular}{l|ccccc}
    \hline
         \textbf{Method} & \multicolumn{3}{c}{$F1@\{10,25,50\}$} & Edit & MoF\\\hline
         Supervised & 30.5 & 25.4 & 17.3 & 26.3 & 43.1 \\
         ICC-wo-unsupervised & 42.6	& 37.5 & 25.3 & 35.2 & 53.4 \\
         ICC-with-unsupervised & \textbf{52.9} & \textbf{49.0} & \textbf{36.6} & \textbf{45.6} & \textbf{61.3} \\
    \hline
    \end{tabular}
    }\end{center}
    \caption{\textbf{``ICC-wo-unsupervised''} (removing the initial unsupervised representation learning from ICC) on 50salads with 5\% $\calD_L$. The ICC results are from fourth iteration \ie ($\text{ICC}_4$).}
    \label{tab:wo_unsuper_pretrain}
\end{table}

\begin{table}[t]
\begin{center}
\small{
\begin{tabular}{l|ccccc}
    \hline
    %& \multicolumn{5}{c}{} \\\hline
     & F1@10 & F1@25 & F1@50 & Edit & MoF \\
    \hline
    Unsupervised & 40.8 & 36.2 & 28.1 & 32.4 & 62.5 \\
    $\text{ICC}_\text{2}$ & 51.3 & 46.6 & 36.5 & 44.7 & 61.3 \\
    $\text{ICC}_\text{3}$ & 52.5 & 47.2 & 36.5 & 45.4 & 62.1 \\
    $\text{ICC}_\text{4}$ & 52.6 & 47.7 & 38.1 & 46.7 & 61.3 \\
    \hline
    \end{tabular}
}\end{center}
\caption{\textbf{Improvement in representation} on 50salads for 5\% labelled data with more iterations of ICC. Note: Representation is evaluated with 100\% data with simple Linear Classifier as discussed in section 4.4.}
\label{tab:50Salads_reprentation_learning}
\end{table}

\begin{figure*}
    \centering
    \includegraphics[width=0.9\linewidth]{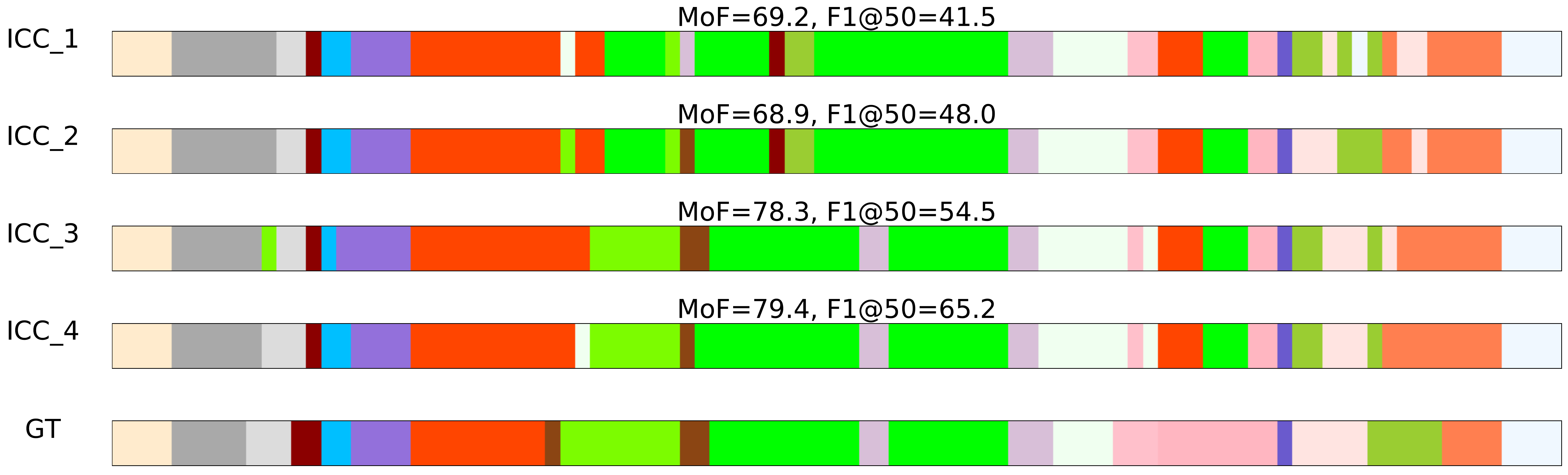}
    \caption{A qualitative example taken from of 50salads, showing progressive improvement in segmentation results with number of iterations of ICC. Some segments becomes more aligned to ground truth(GT) leading improved MoF and F1@50 scores.}
    \label{fig:50salads_icc_example}
\end{figure*}

\begin{table*}[t]
\begin{center}
\small{
% \begin{tabular}{p{0.7cm}|p{1.5cm}|p{0.5cm}p{0.5cm}p{0.5cm}p{0.5cm}p{0.6cm} | p{0.5cm}p{0.5cm}p{0.5cm}p{0.5cm}p{0.6cm} | p{0.5cm}p{0.6cm}p{0.6cm}p{0.5cm}p{0.6cm}}
\begin{tabular}{c|c|ccccc|ccccc|ccccc}
\hline
& & \multicolumn{5}{c|}{Breakfast} & \multicolumn{5}{c|}{50Salads} & \multicolumn{5}{c}{GTEA} \\
\hline
\%$D_L$ &\textbf{Method} & \multicolumn{3}{c}{$F1@\{10,25,50\}$} & Edit & MoF & \multicolumn{3}{c}{$F1@\{10,25,50\}$} & Edit & MoF & \multicolumn{3}{c}{$F1@\{10,25,50\}$} & Edit & MoF \\
\hline
\rowcolor{LightOrange} 
& $\text{ICC}_1$                     
                    & 57.0 & 51.9 & 36.3 & 56.3 & 65.7 
                    & 51.1 & 45.6 & 34.5 & 42.8 & 65.3 
                    & 82.2 & 78.9 & 63.8 & 75.6 & 72.2\\
\rowcolor{LightOrange} 
& $\text{ICC}_2$ 
                    & 60.0 & 54.5 & 38.8 & 59.5 & 66.7
                    & 56.5 & 51.6 & 39.2 & 48.9 & 67.1 
                    & 83.4 & 80.1 & 64.2 & 75.9 & 72.9 \\
\rowcolor{LightOrange} 
& $\text{ICC}_3$
                    & 62.3 & 56.5 & 40.4 & 60.6 & 67.8 
                    & 60.7 & 56.9 & 45.0 & 52.4 & 68.2
                    & 83.5 & 80.8 & 64.5 & 76.3 & 73.1\\
\rowcolor{LightOrange} 
\multirow{-4}{*}{\textbf{$\approx$10}} & $\text{ICC}_4$ 
                    & 64.6 & 59.0 & 42.2 & 61.9 & 68.8
                    & 67.3 & 64.9 & 49.2 & 56.9 & 68.6 
                    & 83.7 & 81.9 & 66.6 & 76.4 & 73.3 \\
                    
\cline{2-17}
& Gain & 7.6 & 7.1 & 5.9 & 5.6 & 3.1 
       & 16.2 & 19.3 & 14.7 & 14.1 & 3.3 
       & 1.5 & 3.0 & 2.8 & 0.8 & 1.1 \\
\hline
\rowcolor{LightGreen} 
& $\text{ICC}_1$ 
                & 68.1 & 63.5 & 49.4 & 66.5 & 72.2
                & 72.2 & 69.1 & 59.9 & 62.2 & 79.7 
                & 89.9 & 88.1 & 79.2 & 84.7 & 80.9\\
\rowcolor{LightGreen}
& $\text{ICC}_2$
                & 71.5 & 67.6 & 54.5 & 68.0 & 74.9 
                & 78.4 & 75.8 & 68.5 & 71.1 & 82.9
                & 89.9 & 88.3 & 75.6 & 85.3 & 80.9\\
\rowcolor{LightGreen}
& $\text{ICC}_3$
                & 71.9 & 68.2 & 54.9 & 68.5 & 75.0 
                & 81.3 & 79.3 & 71.9 & 74.4 & 84.5
                & 90.1 & 88.3 & 78.3 & 86.8 & 81.0 \\
\rowcolor{LightGreen}
\multirow{-4}{*}{\textbf{100}} & $\text{ICC}_4$ 
                & 72.4 & 68.5 & 54.9 & 68.6 & 75.2
                & 83.8 & 82.0 & 74.3 & 76.1 & 85.0 
                & 91.4 & 89.1 & 80.5 & 87.8 & 82.0 \\
\cline{2-17}
& Gain & 4.3 & 5.0 & 5.5 & 2.1 & 3.0 
       & 11.6 & 12.9 & 14.4 & 13.9 & 5.3
       & 1.5 & 1.0 & 1.3 & 3.1 & 1.1 \\
\hline

\end{tabular}
}\end{center}
\caption{Quantitative evaluation of progressive semi-supervised improvement with more iterations of \textbf{ICC with $\approx$ 10\% and 100\% labelled data}.}
\label{tab:icc_10_100}
\end{table*}

\subsubsection{Unsupervised Clusterring vs Representation Learning:} We also report the linear classification scores for the publicly available checkpoints of unsupervised segmentation CTE-MLP~\cite{unsupervised-kukleva2019unsupervised} (which pre-trains representation for predicting the absolute temporal positions of features) in Table~\ref{tab:unsupervised_sota}. Surprisingly, linear classifier accuracy of the representation from there unsupervised work is below the input IDT features baseline for 50Salads. This is likely due to there assumptions made of embedding the absolute temporal positions to the features before clustering, when positions of actions within 50salads is widely different in different videos of the dataset. However, we note, as discussed in related work section 2 of the main paper, unsupervised works like ~\cite{unsupervised-kukleva2019unsupervised, unsupervised-li2021action, sarfraz2021temporally} is based on development of clustering algorithms, with requirement of higher order viterbi algorithm. Using the same representation with two different clustering algorithm, the segmentation results can vary widely. For example, IDT features of 50salads has MoF of 29.7\% with Viterbi+Kmeans and 66.5\% with TWFINCH~\cite{sarfraz2021temporally} as shown in ~\cite{sarfraz2021temporally}. So our unsupervised representation learning step is not directly comparable to unsupervised clustering(segmentation) algorithms. They can only evaluate there unsupervised clusters based on Hungarian Matching to ground truth labels (\ie no classifier is trained, only creates segmentation without labelling and  temporal action segmentation requires to jointly segment and classify all actions) and therefore they cannot be compared to fully-semi-weakly supervised works.

\subsection{ICC without unsupervised step.}\label{subsec:icc_wo_pretraining}

In table~\ref{tab:wo_unsuper_pretrain}, we show results of our ICC without the initial \textit{``unsupervised representation learning''} as \textbf{``ICC-Wo-Unsupervised''}. This essentially means that the $2^{nd}$ row of table~\ref{tab:wo_unsuper_pretrain} represents the scenario in which from our ICC algorithm we remove the $1^{st}$ contrast step that is learned with cluster labels. The improvement in scores over supervised setup is quite low compared to the full-ICC \textit{with} the unsupervised pre-training, shown as \textbf{``ICC-With-Unsupervised''}. This verifies the importance of our unsupervised learning step in ICC.

\subsection{Iterative progression of ICC results} \label{subsec:icc_progression_results}

We discuss our detailed semi-supervised algorithm in section 5 of our main paper and provide visualization of the algorithm in Figure 4. We summarize it with an Algorithm~\ref{algo:contrast_classify} in supplementary.

\begin{table*}[t!]

\begin{center}
\small{
\begin{tabular}{l|l|ccccc}
    \hline
     Dataset & ICC(Num Videos) & F1@10 & F1@25 & F1@50 & Edit & MoF \\
    \hline
    \rowcolor{LightCyan}
    &  $\text{ICC}_1$ ($\approx 63$ Videos) & 54.5 $\pm$  1.2 & 48.7 $\pm$ 1.1  & 33.3 $\pm$ 1.1 & 54.6 $\pm$ 0.9 & 64.2 $\pm$  1.3 \\
    \rowcolor{LightCyan}
    & $\text{ICC}_4$ ($\approx 63$ videos) &  60.2 $\pm$ 1.5 & 53.5 $\pm$ 1.3 & 35.6 $\pm$ 0.9 & 56.6 $\pm$ 1.2 & 65.3 $\pm$ 1.8 \\
    \cline{2-7}
    \rowcolor{LightCyan} 
    & $\text{ICC}_1$ ($\approx 120$ Videos) & 57.0 $\pm$ 1.9 & 51.9 $\pm$  2.1 & 36.3 $\pm$ 1.3 & 56.3 $\pm$ 1.2 & 65.7 $\pm$ 1.9 \\
    \rowcolor{LightCyan} 
    \multirow{-4}{*}{Breakfast}  & $\text{ICC}_4$ ($\approx 120$ Videos) &  64.6 $\pm$ 2.1 & 59.0 $\pm$ 1.9 & 42.2 $\pm$ 2.5 & 61.9 $\pm$ 2.2 & 68.8 $\pm$ 1.3 \\\hline
    
    \rowcolor{LightPink}
    & $\text{ICC}_1$ (3 Videos) & 41.3 $\pm$ 1.9 & 37.2 $\pm$ 1.5 & 27.8 $\pm$ 1.1 & 35.4 $\pm$ 1.6 & 57.3 $\pm$ 2.3 \\
    \rowcolor{LightPink}  
    & $\text{ICC}_4$ (3 videos) &  52.9 $\pm$ 2.2 & 49.0 $\pm$ 2.2 & 36.6 $\pm$ 2.0 & 45.6 $\pm$ 1.4 & 61.3 $\pm$ 2.3 \\
    \cline{2-7}
    \rowcolor{LightPink} 
    & $\text{ICC}_1$ (5 Videos) & 51.1 $\pm$  2.1 & 45.6 $\pm$  1.3 & 34.5 $\pm$ 1.7 & 42.8 $\pm$ 1.1 & 65.3 $\pm$  0.8 \\
    \rowcolor{LightPink} 
    \multirow{-4}{*}{50salads} & $\text{ICC}_4$  (5 videos) & 67.3 $\pm$ 1.8 & 64.9 $\pm$ 2.5 & 49.2 $\pm$ 1.8 & 56.9 $\pm$ 2.1 & 68.6 $\pm$  0.7 \\
    \hline
    \end{tabular}
}\end{center}
\caption{\textbf{Mean and standard deviation for 5 different selections} of 5\% and 10\% labelled videos from Breakfast and 50salads. For each metric we report the results in the format $mean \pm std$, \ie the means and the standard deviation for the $5$ runs.}
\label{tab:50Salads_error_bars_5}
\end{table*}

\subsubsection{Improvement after \textit{Contrast} step}
In Table \ref{tab:50Salads_reprentation_learning} we show the improvement in representation after each \textit{contrast} step. Due to the usage of better pseudo-labels obtained from the preceding \textit{classify} step, the following contrast step results in better representations as more iteration is performed. For 5\% labelled videos of the 50salads dataset, we can see that there is a clear improvement in F1 and Edit scores as more iterations are performed. Note that the evaluation of the learned representation is same as described in subsection 4.4 of the main text.

\subsubsection{Improvement after \textit{Classify} step}
In Table 2 of main text, we showed the progressive improvement in performance for 5\% labelled videos, evaluated after the \textit{classify} step of each ICC iteration. In Table~\ref{tab:icc_10_100} we show the same progressive improvements for 10\% and 100\% labelled videos. The evaluation is done after the \textit{classify} step of each iteration of the algorithm.Our ICC raises overall scores on all datasets, with stronger improvements in F1 and Edit scores. 

\textbf{Qualitative visualization of segmentation} We show in Figure~\ref{fig:50salads_icc_example} an example segmentation results from 50salads dataset of how the segmentation results improves (becomes more aligned to GT shown with increase in MoF and F1@50) with more iterations of ICC.   

\subsection{Standard deviations in results} \label{subsec:mean_std_results} We show our standard deviations in results for 50Salads, Breakfast dataset for variations in labelled data used in Table~\ref{tab:50Salads_error_bars_5}. We show the variation in results for $\text{ICC}_1$ and $\text{ICC}_4$ when we take 5 different random selections of 5\%, 10\% labelled videos in Breakfast and 50Salads from corresponding training splits. We report the mean and standard deviation for different choices as $mean \pm std$ format.

\bibliography{aaai22.bib}
\end{document}

\maketitle

In section~\ref{sec:implementation_details} we details our training hyper-parameters used for unsupervised and semi-supervised setup. We also provide variations in results with change in important hyper-parameters of our algorithm to experimentally validate our choice of hyper-parameters. In section~\ref{sec:base_architecture} we provide additional detail of the base architecture C2F-TCN~\cite{singhania2021coarse} used and validate the choice with respect to other base model like ED-TCN~\cite{TED-lea2017temporal}, MSTCN~\cite{li2020ms}. In section~\ref{sec:multi_resolution_feature} we experimentally validate the normalization strategy used for our multi-resolution feature and show derivation of temporal continuity implicitly incorporated in it. In section~\ref{sec:additional_ablation} we show additional experimental and qualitative visualization results. We show visualization of representation learnt and input I3D feature, and show results with IDT as input features in section~\ref{subsec:IDT}. The importance of our unsupervised representation learning step in final results is shown in section~\ref{subsec:icc_wo_pretraining}. We show quantitative and qualitative improvements in representations learnt and semi-supervised classification results with multiple iteration of ICC in section ~\ref{subsec:icc_progression_results}. Finally we report deviations of our results with different selections of labelled videos in ~\ref{subsec:mean_std_results}.

\section{Implementation Details \& Hyperparameter Selection}\label{sec:implementation_details}

% ~\AY{state somewhere that you are choosing your hyperparameters on a validation set and tuned on test set directly.}
% For the input clustering algorithm of input features described in section 4.1 of the main paper, we use the standard $K$-Means clustering algorithm. 
All our experiments are conducted on \textit{Nvidia GeForce RTX 2080 Ti} GPUs with 10.76 Gb memory. We use the Adam optimizer with Learning Rate (LR), Weight Decay (WD), Epochs (Eps), and Batch Size (BS) used for our unsupervised and semi-supervised setup shown in Table \ref{tab:hyperparameters}. For the \textit{contrast step} of $\text{ICC}_i$ for $i \ge 2$, we reduce the learning rate to be 0.1 times that of the \textit{contrast step} learning rate used in the first iteration (\ie $\text{ICC}_1$). We use a higher number of epochs for the semi-supervised fine-tuning step, but as the number of labeled data is minimal, the training time is quite less. For all three datasets, we follow the $k$-fold cross-validation averaging to report our final results of representation, semi-supervised and fully-supervised results(however, we tune our model hyper-parameters for a hold-out validation set from training dataset). Here $k=\{4, 5, 4\}$ for Breakfast, 50Salads and GTEA respectively and is same as used in other works ~\cite{li2020ms, selfsupervised-chen2020action}. 
% The $k$-fold cross-validation splits of Breakfast, 50Salads, and GTEA consists of $k=\{4, 5, 4\}$ respectively as used in ~\cite{li2020ms}. 
% with higher number epochs.

\subsection{Sampling strategy, number of samples $2K$} We show ablations for the choice of $2K$ \ie number of representation samples drawn per video as described in section 4.1 of main paper in Table~\ref{tab:num_samples}. Thus our value of $2K$ is determined by experimental validation. In 50salads with $2K=120$ and batch size of 50, we get roughly 0.6 million positive samples per batch, with each positive sample having roughly around 6.5K negative samples. 

\begin{table}[t]
    \begin{center}
        \small{
    
    \begin{tabular}{l|ccccc}
    \hline
         Samples($2K$) & \multicolumn{3}{c}{$F1\{10, 25,50\}$} & Edit & MoF \\\hline
         60 & 38.5 & 33.2 & 25.1 & 29.9 & 62.5\\
         120 & 40.8	& 36.2 & 28.1 & 32.4 & 62.5\\
         180 & 39.1	& 34.5 & 27.3 & 30.8 & 62.1\\\hline
    \end{tabular}
    }\end{center}
    \caption{Ablation results of \textbf{number of samples per video} required for representation learning.}
    \label{tab:num_samples}
\end{table}

\begin{table}[]
    \begin{center}
    \small{
    \begin{tabular}{l|l|ccccc}
    \hline
     \%$\calD_L$ & \textbf{Method} & \multicolumn{3}{c}{$F1@\{10,25,50\}$} & Edit & MoF\\\hline
        100\% & Full-Supervised & 68.0 & 63.9 & 52.6 & 52.6 & 64.7 \\
        \hline
         5\% & Supervised &  32.4 & 26.5 & 14.8 & 25.5 & 39.1 \\
         5\% & our ICC &  39.3 & 34.4 & 21.6 & 32.7 & 46.4 \\
         \cline{2-7}
         5\% & Gain & 6.9 & 7.9 & 6.8 & 7.2 & 7.3 \\
         \hline
    \end{tabular}
    }\end{center}
    \caption{\textbf{Our semi-supervised (final $\text{ICC}_4$) results with EDTCN} \cite{TED-lea2017temporal} on 50salads with 5\% labelled data significantly improves over its supervised counterpart.}\label{tab:EDTCN_semi_super}
\end{table}

\begin{table*}[]
\begin{center}
    \small{
    % \begin{tabular}{p{2.4cm}|p{0.6cm}p{0.6cm}p{0.4cm}p{0.3cm}| p{0.6cm}p{0.6cm}p{0.4cm}p{0.3cm}|p{0.6cm}p{0.6cm}p{0.4cm}p{0.3cm}}
    \begin{tabular}{c|cccc|cccc|cccc}
    \hline
        & \multicolumn{4}{c|}{Breakfast} 
        & \multicolumn{4}{c|}{50Salads} 
        & \multicolumn{4}{c}{GTEA}\\
        \cline{2-13}
        Algorithm & LR & WD & Eps. & BS 
                  & LR & WD & Eps. & BS
                  & LR & WD & Eps. & BS\\
        \hline
        \textit{Contrast step} (model $\mathbf{M}$) & 1e-3 & 3e-3 & 100 & 100 
                     & 1e-3 & 1e-3 & 100 & 50 
                     & 1e-3 & 3e-4 & 100 & 30 \\
        \hline
        \textit{Classify step} (classifier $\bG$) & 1e-2 & 3e-3 & 700 & 100
                            & 1e-2 & 1e-3 & 1800 & 5 
                            & 1e-2 & 3e-4 & 1800 & 5 \\
        \textit{Classify step} (model $\mathbf{M}$) & 1e-5 & 3e-3 & 700 & 100
                                  & 1e-5 & 1e-3 & 1800 & 5 
                                  & 1e-5 & 3e-4 & 1800 & 5 \\
    \hline
    \end{tabular}
    }\end{center}
    \caption{Training hyperparameters learning rate(LR), weight-decay(WD), epochs(Eps.) and Batch Size(BS) used for different datasets for unsupervised and semi-supervised learning.}\label{tab:hyperparameters}
\end{table*}

%We sample frames from each video with $K=60$ partitions, {$\epsilon\!\!\approx\!\!\frac{1}{3K}$=0.005} for sampling, and temporal proximity $\delta\!=\!0.5$ for 50Salads.  
%These values are roughly aligned with the average length of action in the dataset. 
% The contrastive temperature $\tau$ in Eqs.~\eqref{eqn:frame-level-contrast} and \eqref{eqn:video-level-contrast} is set to $0.1$. We also leverage the feature augmentations of C2F-TCN,

% Formally, for $1\le i\le K$ and with a chosen $\epsilon < 1/K$, we sample $t^n_i \sim \text{unif}(\frac{i-1}{K}, \frac{i}{K} - \epsilon)$ and set $t^n_{i+K} = t^n_{i} + \epsilon$. ~\AY{if space, illustrate sampling procedure with diagram}

\begin{table}[]
    \begin{center}
    \small{
    \begin{tabular}{cccc}
        \hline
         Cluster-Type (Num-Clusters) & Breakfast & 50Salads & GTEA \\
         \hline
         FINCH ($A$)  & 61.7 & 56.3 & 56.7 \\
         Kmeans ($A$) & 70.0 & 60.4 & 65.6 \\
         \textbf{Kmeans ($\approx 2A$)} & \textbf{70.5} & \textbf{62.5} & \textbf{69.1} \\
        \hline
    \end{tabular}
    }\end{center}
    \caption{Unsupervised Representation's MoF variation with different clustering types and number of clusters used during training. $A$ denotes number of unique actions in the dataset.}
    \label{tab:kmeans_finch}
\end{table}

\subsection{Input feature clustering} 
In section 4.1 of the main paper, unsupervised feature learning requires cluster labels from the input features. %For the clustering algorithm, we 
We cluster at the mini-batch level with a standard {$k$-means} and then compare with % algorithm. 
Finch~\cite{sarfraz2019efficient}, an agglomerative clustering, that have shown to be useful in the unsupervised temporal segmentation \cite{sarfraz2021temporally}. %So we also consider using the Finch clustering and compare it to $K$-means in
Comparing the two in Table~\ref{tab:kmeans_finch}, we observe that %find that . We empirically find 
$K$-means performs better.  We speculate that this is because %to be working better than Finch, intuitively because 
Finch is designed for %shown to be beneficial for 
per-video clustering. In contrast, our clustering on the mini-batch is on a dataset level, i.e., over multiple video sequences of different complex activities. %and we do an overall dataset level clustering. 

For choosing $k$ in the $k$-Means clustering, we choose $\approx 2A$ ($A$ denotes number of unique actions) number of clusters resulting in $K=\{100, 40, 30\}$ for Breakfast, 50Salads, and GTEA datasets, respectively.  The advantage of using  $\approx 2A$ clusters versus simply $A$ is verified in Table~\ref{tab:kmeans_finch}. The improvement is greater for datasets with fewer action classes like GTEA and 50salads than the Breakfast action dataset.
% ~\AY{could comment on why 2A does not improve much for Breakfast but more so for the smaller datasets}

\section{Details And Choice Of Base Architecture}\label{sec:base_architecture}
\subsection{Details of C2F-TCN}
Our base architecture is same as C2F-TCN~\cite{singhania2021coarse}, which takes snippet-level features like IDT or I3D as input and which has convolution block used for encoder-decoder of kernel size $3$ with total of $4.07$ million trainable parameters. To train the C2F-TCN, we use %\textit
{downsampled} input features and a %\textit
{multi-resolution augmentation} training strategy as recommended by \cite{singhania2021coarse}. We briefly discuss downsampling and augmentation used here and also show its effect on our unsupervised feature learning in Table \ref{tab:feature-agument}.

\textbf{Downsampling:}
The pre-trained input feature representations i.e I3D and IDT referenced as $\vid \in \bbR^{T \times F}$ and the ground truth $y$ of temporal dimension $T$ is downsampled by $w$ to obtain the input feature vector $\vid_{in} \in \bbR^{T_{in} \times F}$ and ground truth $y_{in}$ of temporal dimension $T_{in}$, where $T_{in}= \frac{T}{w} $. This is done by max-pooling the features using a temporal window of size $w$ and assigning the corresponding label which is most frequent in the given window.  Specifically, if the input feature and ground truth label at time $t$ i.e $\vid^{w}_{in}[t]$ and $y^{w}_{in}[t]$ for some temporal window $w > 0$ is, then the downsampled feature and label can be defined as 
\begin{align}\label{eqn:downsample}
    \vid^{w}_{in}[t] &= \max_{\tau \in \left[wt, wt+w\right)} \vid[\tau] \\ 
    y^{w}_{in}[t] &= \underset{k \in \calA}{\argmax} \sum_{\tau=wt}^{wt+w} \bbI\sqBK{y\left[\tau\right]=k}.
\end{align}

\textbf{Temporal feature augmentation} applies max-pooling %with a fixed window size, the max-pooling is done using 
over multiple temporal resolution of the features by varying the window size $w$ uniformly from $[w_{min}, w_{max}]$ during training.
% During test time a base window $w_0$ is chosen and all outputs are upsampled to original input video temporal dimension $T$ to obtain the final scores.
We use the same values of $w_0$ as~\cite{singhania2021coarse} with $w_0\!=\!\{10, 20, 4\}$ for Breakfast, 50Salads and GTEA respectively and $w_{min}\!=\!\frac{1}{2} w_0$   and $w_{max}\!=\!2 \times w_0$. In Table~\ref{tab:feature-agument} we show the improvements in unsupervised features linear evaluation scores when training with feature-augmentation. We do not use test time feature augmentations and all results in our paper is reported without test time augmentations.

\begin{table*}[t]

\begin{center}
\small{
\begin{tabular}{l|ccccc|ccccc|ccccc}
\hline
& \multicolumn{5}{c|}{Breakfast} 
& \multicolumn{5}{c|}{50Salads} 
& \multicolumn{5}{c}{GTEA}\\
\hline
\textbf{Method} &
\multicolumn{3}{c}{$F1\{10, 25,50\}$} & Edit & MoF & \multicolumn{3}{c}{$F1\{10, 25,50\}$} & Edit & MoF &
\multicolumn{3}{c}{$F1\{10, 25,50\}$} & Edit & MoF \\
\hline
No Augment & 55.6 & 50.2 & 36.5 & 49.4 & 69.4 & 
              40.0 & 34.1 & 27.0 & 31.0 & 62.3 &
              70.0 & 63.4 & 47.2 & 65.6	& 69.0 \\
Augment & \textbf{57.0} & \textbf{51.7} & \textbf{39.1} & \textbf{51.3} & \textbf{70.5} & 
 \textbf{40.8} & \textbf{36.2} & \textbf{28.1} & \textbf{32.4} & \textbf{62.5} &
 \textbf{70.8} & \textbf{65.0} & \textbf{48.0} & \textbf{65.7} & \textbf{69.1} \\
 
% No Augment & 50.2 & 36.5 & 49.4 & 69.4 & 
%              34.1 & 27.0 & 31.0 & 62.3 &
%              63.4 & 47.2 & 65.6	& 69.0 \\
% Augmentation & 51.7 & 39.1 & 51.3 & 70.5 & 
%           36.2 & 28.1 & 32.4 & 62.5 &
%           65.0 & 48.0 & 65.7 & 69.1 \\

\hline
\end{tabular}
}\end{center}
\caption{\textbf{Impact of using Multi-Resolution Augmentation}} \label{tab:feature-agument}
\end{table*}

\begin{table*}[h]

 \begin{center}
    \small{
\begin{tabular}{l|ccccc|ccccc|ccccc}
\hline
& \multicolumn{5}{c|}{Breakfast} 
& \multicolumn{5}{c|}{50Salads} 
& \multicolumn{5}{c}{GTEA}\\
\hline
\textbf{Method} &
\multicolumn{3}{c}{$F1\{10, 25,50\}$} & Edit & MoF & \multicolumn{3}{c}{$F1\{10, 25,50\}$} & Edit & MoF &
\multicolumn{3}{c}{$F1\{10, 25,50\}$} & Edit & MoF \\
\hline

Alternate $\feat^{'}[t]$ &44.3 & 38.3 & 26.1 & 40.9 & 60.9 &
                          32.9 & 27.3 & 19.9 & 26.5 & 51.2 &
                          56.4 & 48.6 & 31.3 & 52.1 & 58.9\\
\textbf{Proposed} $\feat[t]$ & \textbf{57.0} & \textbf{51.7} & \textbf{39.1} & \textbf{51.3} & \textbf{70.5} & 
                     \textbf{40.8} & \textbf{36.2} & \textbf{28.1} & \textbf{32.4} & \textbf{62.5} &
                     \textbf{70.8} & \textbf{65.0} & \textbf{48.0} & \textbf{65.7} & \textbf{69.1} \\
% Naive $\feat^{'}[t]$ & 38.3 & 26.1 & 40.9 & 60.9
%                          & 27.3 & 19.9 & 26.5 & 51.2
%                          & 48.6 & 31.3 & 52.1 & 58.9\\
% Proposed $\feat[t]$ & 51.7 & 39.1 & 51.3 & 70.5 &  
%                      36.2 & 28.1 & 32.4 & 62.5 &
%                      65.0 & 48.0 & 65.7 & 69.1 \\

\hline
\end{tabular}
}\end{center}
\caption{\textbf{Importance of normalization order for Multi-Resolution Feature}}\label{tab:multi-res-feature} \label{tab:feature_similarity_formation}
\end{table*}

\begin{algorithm}[t]
\caption{Iterative \textit{Contrast-Classify} algorithm}\label{algo:contrast-classify}
\begin{algorithmic}[1]
% \Comment{\emph{Contrast-classify}}
\While {\textit{iter} $\le$ MaxIter}
    \For {\emph{epoch} $\le$ MaxEpoch} \Comment{\textbf{\emph{Contrast}}:}
    \State sample minibatch $\{\vid_n\}_{n=1}^N \subset \calD_u \cup \calD_l$
        \If {\emph{iter} $= 0$}
            \State $\{l_n\}_{n=1}^N \gets $ \emph{Cluster}$\big(\{\vid_n\}_{n=1}^N\big)$
        \EndIf
        \State $\losscons \gets$ \emph{Contrastive}$\big(\{\BK{\vid_n, l_n}\}_{n=1}^N, \mathbf{M}\big)$
        \State minimize $\losscons$ and update $\mathbf{M}$
    \EndFor
    \For {$\vid_n \in \calD_l$}
        \State $l_n \gets y_n$
    \EndFor
    \For {\emph{epoch} $\le$ MaxEpoch}\Comment{\textbf{\emph{Classify}}:} 
        \State sample minibatch $\{\vid_n\}_{n=1}^N \subset \calD_l$
        \State $\losscons \gets$ \emph{Contrastive}$\big(\{\BK{\vid_n, l_n}\}_{n=1}^N, \mathbf{M}\big)$
        \State $\{\bar{\bp}_n\}_{n=1}^N \gets$ \emph{Predict}$(\{\vid_n\}_{n=1}^N, \bG)$
        \State $\losscross \gets$ \emph{CrossEntropy}$\big(\{\BK{\bar{\bp}_n, y_n}\}_{n=1}^N\big)$
        \State $\calL \gets \losscons + \losscross$
        \State minimize $\calL$ and update $\mathbf{M}$ and $\bG$
    \EndFor
    \For {$\vid_n \in \calD_u$}
        \State $l_n \gets$ \emph{Predict}$(\vid_n, \bG)$
    \EndFor
    \State \emph{iter} $\gets$ \emph{iter} $+ 1$
\EndWhile
\end{algorithmic}\label{algo:contrast_classify}
\end{algorithm}

\subsection{Other baseline models}

% ~\AY{for both training and testing}. 

% \begin{figure*}
%   \includegraphics[trim={3cm 3cm 4cm 2cm}, width=0.9\linewidth]{images/contrast-classify.pdf}
%   \caption{\textit{Iterative-Classify-Contrast} algorithm and illustration}
%   \label{alg.semi-sup}
% \end{figure*}

% ~\AY{introduce that you try the entire algorithm with 2 other baselines.  start with ED-TCN, then move onto MS-TCN. bring table closer here to text.}

We try our entire algorithm~\ref{algo:contrast_classify} for semi-supervised temporal segmentation (outlined in section 5, Figure 4 of main paper) with unsupervised representation learning and iterative contrast classify with two other base TCN models: ED-TCN(an encoder-decoder architecture) and MSTCN (wavenet like refinement architecture).

\textbf{ED-TCN:} We show our proposed ICC to be working with ED-TCN~\cite{TED-lea2017temporal} in Table~\ref{tab:EDTCN_semi_super}, where ICC improves performance over its supervised counterpart. Due to smaller capacity of ED-TCN compared to C2F-TCN (indicated by the fully supervised performance (top row of table~\ref{tab:EDTCN_semi_super}) of ED-TCN 64.7\% vs C2F-TCN 79.4\% MoF), ICC algorithm improvement with ED-TCN is lower than improvement with C2F-TCN. For constrastive framework, an increased  performance with better model capacity has also been shown before in SimCLR~\cite{simCLR}.

\textbf{MSTCN:} Our proposed algorithm does not work well with MSTCN~\cite{li2020ms} architecture, possibly due to the fact that MSTCN is not designed for representation learning as 3 out of the 4 model blocks consists of refinement stages, where each stage takes \textit{class probability} vectors as input from the previous stages. Therefore, representation learning and classifier cannot be decoupled, making alternative classifier-representation learning algorithm impossible. Further, MS-TCN do not have multiple temporal resolution representation like encoder-decoder architecture which play a significant role in our contrastive learning as discussed earlier. 
\begin{table}[t]
    \begin{center}
    \small{
    \begin{tabular}{lcc}
    \hline
    \textbf{Representation} & \textbf{Breakfast} & \textbf{50Salads} \\
    \hline
    % IDT & GMM\cite{unsupervised-sener2018unsupervised} & 34.6 & - \\
    % IDT + CTE-MLP\cite{unsupervised-kukleva2019unsupervised} &  $k$-means + Viterbi + Hungarian @Activity & 41.8 & 30.2 \\
    % IDT + VT-UNET\cite{unsupervised-vidalmata2021joint} & $k$-means + Viterbi + Hungarian @Activity & 48.1 & 24.2 \\
    % IDT + ASAL\cite{unsupervised-li2021action} & $k$-means + Viterbi + Hungarian @Activity & 52.5 & 34.4 \\
    % IDT & $k$-means + Hungarian @Video & 42.7 & 29.7 \\
    % IDT & TWFinch \cite{sarfraz2021temporally} + Hungarian @Video & 62.7 & 66.5 \\
    % \midrule
    IDT + CTE-MLP & 45.0$^*$ & 34.1$^*$ \\\hline
    
    IDT & 18.2 & 36.9 \\
    
    % IDT + CTE-MLP\cite{unsupervised-kukleva2019unsupervised} & Linear classifier, $k$-fold @Activity & 51.2$^*$ & - \\
    \textbf{Ours} (Input IDT) & 67.9 & 52.2 \\
    \hline
    Gain  & +49.7 & +15.3 \\
    % IDT + \textbf{Ours} &  68.3 & - \\
    \hline
    \end{tabular}
    }\end{center}
    \caption{Our learned representation linear evaluation of segmentation's  MoF significantly improves upon \textbf{input IDT features}. $^*$ indicates evaluations based on publicly available checkpoints.}
    \label{tab:unsupervised_sota}
\end{table}

\section{Multi-Resolution Features} \label{sec:multi_resolution_feature}
\textbf{Normalization:}
% In Section 3.4 of the main paper, we outlined our normalization procedure when forming the multi-resolution feature.  % we introduce the proposed way to calculate form the multi-resolution similarity feature, with normalization before concatenation. 
Our proposed multi-resolution feature, as outlined in Section 4.3 of the main paper, is defined for frame $t$ as $\feat[t] = \BK{\bar{\bz}_1[t]:\bar{\bz}_2[t]:\ldots:\bar{\bz}_6[t]}$, where $\bar{\bz}_u[t] = {\hat{\bz}}_u[t]/\norm{{\hat{\bz}}_u[t]}$, \ie ${\hat{\bz}}_u[t]$, the upsampled feature from decoder $u$, is normalized first for each frame and then concatenated along the latent dimension. 
An alternative and naive construction would be to apply normalization after concatenation,~\ie{} $\feat^{'}[t] = \BK{\hat{\bz}_1[t]:\hat{\bz}_2[t]:\ldots:\hat{\bz}_6[t]}$. The features $\hat{\bz}_u$ are the upsampled \textit{un-normalized} feature vector of decoder layer $u$. Note that a final normalization of $\feat^{'}[t]$ is no longer necessary as the cosine similarity is invariant.  We verify in \ref{tab:feature_similarity_formation} that applying normalization \textit{before} concatenation is critical.  %before concatenation.

% refers to the alternate feature 
% $\feat^{''} = \feat^{'}[t]/\norm{\feat^{'}[t]}$, where

% As explained in section 3.4, significant difference in performance can be seen in 

\paragraph{Inherent temporal continuity of our feature $\feat$} The inherent temporal continuity encoded in our multi-resolution feature is discussed in Section 4.3 of the main paper.  For the `\textit{nearest}' neighbor upsampling strategy, the multi-resolution feature $\feat$ has the property of being similar for nearby frames. Coarser features like $\{\bz_1, \bz_2, \bz_3\}$ are more similar than fine-grained features at higher decoder layers. This also gives independence to the higher resolution features to have high variability even for nearby frames.
%~\AY{state property in words before defining mathematically}%,  we can have the following property of the multi-resolution feature $\feat$. 
Specifically, for two frames $t, s \in \bbN$, if $\floor*{t/2^u} = \floor*{s/2^u}$ for some integer $u > 0$ then $\simi{\feat[t]}{\feat[s]} \ge 1 - u/3$. This follows from the fact that for nearest upsampling, $\floor*{t/2^u} = \floor*{s/2^u}$ for some $0 \le u \le 5$, implies that
\begin{align}\label{eq.nearest_meaning}
    \bz_v[t] = \bz_v[s] \quad\text{for all}\quad 1\le v \le 6-u.
\end{align}
Meaning, the lower resolution features coincide for proximal frames. As discussed and shown in equation (6) of the main text, all the layers have equal contribution while calculating similarity of our multi-resolution feature. For $t,s$, with $\floor*{t/2^u} = \floor*{s/2^u}$ for some $0 \le u \le 5$, we start with the equation (6) of main text to derive --
% ~\AY{one sentence missing here to give some transition to equations?}
\begin{align*}
    & \simi{\feat[t]}{\feat[s]}\\ 
    &= \sum_{v=1}^6 \frac{1}{6} \cdot \simi{\bz_v[t]}{\bz_v[s]}\\
    &= \sum_{v=1}^{6-u} \frac{1}{6} \cdot \simi{\bz_v[t]}{\bz_v[s]} + \sum_{v=7-u}^{6} \frac{1}{6} \cdot \simi{\bz_v[t]}{\bz_v[s]}\\
    &= \frac{6-u}{6} + \sum_{v=7-u}^{6} \frac{1}{6} \cdot \simi{\bz_u[t]}{\bz_u[s]} \quad \BK{\text{from \eqref{eq.nearest_meaning}}}\\
    % &\ge \frac{6-u}{6} + \sum_{v=7-u}^{6} \frac{1}{6} \cdot (-1) \qquad\qquad \BK{\text{as}\,\, \simi{\cdot}{\cdot} \ge -1}\\
    &\ge \frac{6-u}{6} - \frac{u}{6} \qquad\qquad \BK{\text{as}\,\, \simi{\cdot}{\cdot} \ge -1}\\
    &= 1 - \frac{u}{3}.
\end{align*}
That means for $\floor*{t/2^u} = \floor*{s/2^u}$ for some $0 \le u \le 5$ implies $\simi{\feat[t]}{\feat[s]} \ge 1 - \frac{u}{3}$. The inequality is trivial for $u > 5$.

% \section{Additional Algorithmic Details}

% \section{Details of base architecture and augmentation} 
% ~\AY{switch order with unsupervised learning}
% In section 3.1 

% \begin{table}[]
%     \caption{Unsupervised Linear(MF) Scores variation with number clusters used.}
%     \label{tab:kmeans_num_clusters}
%     \centering
%     \begin{tabular}{cccc}
%         \toprule
%          Num Clusters & Breakfast MF & 50Salads MF & GTEA MF\\
%          \midrule
%          $A$ & 70.0 & 60.4 & 65.6 \\
%          $\approx 2A$ & 70.5 & 62.5 & 69.1 \\
%         \bottomrule
%     \end{tabular}
% \end{table}

\section{Additional Ablation}\label{sec:additional_ablation}

\subsection{Unsupervised Representation learning}\label{subsec:IDT}
\begin{figure*}[t]
    \centering
    \includegraphics[width=0.8\linewidth]{images/umap_breakfast.pdf}
    \caption{Change in UMAP scatter plot~\cite{UMAP} after our unsupervised representation learning. The plot is of seven videos of the complex activity ``\textit{fry\_egg}". Each point in the plot denotes a frame of a video while different colors represent different actions (specified in the legend). Left subplot shows the discriminativeness of input I3D features, while the right one shows representation obtained by our unsupervised learning. In the left figure, each of the connected path-like blobs belong to different videos. Meaning the I3D features of the frames of the same video are well-connected in the feature space and suggest that temporal continuity is the primary factor behind the distinction in the feature space. In contrast (in the right figure), our learned representation separates actions across different videos. There are also some locally-connected components in the right figure which actually belong to the frames of the same video which are of the same action as well. Hence, rather than separating frames primarily based on temporal continuity (like the I3D feature) our learned feature primarily separates based on action, and then locally on the basis of temporal continuity (same video).}
    \label{fig:Umap_representation}
\end{figure*}
\subsubsection{Visualization of representations learnt:} In Table 1 of the main paper we show the representations results for I3D features as input to C2F-TCN. In Figure~\ref{fig:Umap_representation} we visualize our learnt representation versus input I3D features with the help of UMAP(used for visualization by reducing data to 2-dimensions) visualization. We show our learnt representation incorporate separability based of actions and therefore have much higher segmentation linear classifier scores that input I3D features(Table 1 of main paper).

\subsubsection{Experiment Result with IDT features:} We omitted representations segmentation results when using IDT features as inputs to C2F-TCN instead of I3D due to space limitations and include it here in Table~\ref{tab:unsupervised_sota}. IDT features are obtained in unsupervised way with use of PCA and Gaussian Mixture Models~\cite{wang2013action} compared to I3D obtained from pre-trained models. We show that trends in improvement of our representation compared to input IDT features are similar to the results of Table 1 of the main paper.

\begin{table}[]
    \begin{center}
    \small{
    \begin{tabular}{l|ccccc}
    \hline
         \textbf{Method} & \multicolumn{3}{c}{$F1@\{10,25,50\}$} & Edit & MoF\\\hline
         Supervised & 30.5 & 25.4 & 17.3 & 26.3 & 43.1 \\
        %  Self-Train-ICC & 40.3 & 35.9 & 25 & 32.5 & 50.7 \\
        %  ICC-wo-multi-resolution & 44.0 & 36.9 & 25.3 &	37.5 & 48.1 \\
         ICC-wo-unsupervised & 42.6	& 37.5 & 25.3 & 35.2 & 53.4 \\
         ICC-with-unsupervised & \textbf{52.9} & \textbf{49.0} & \textbf{36.6} & \textbf{45.6} & \textbf{61.3} \\
    \hline
    \end{tabular}
    }\end{center}
    \caption{\textbf{``ICC-wo-unsupervised''} (removing the initial unsupervised representation learning from ICC) on 50salads with 5\% $\calD_L$. The ICC results are from fourth iteration \ie ($\text{ICC}_4$).}
    \label{tab:wo_unsuper_pretrain}
\end{table}

\begin{table}[t]
\begin{center}
\small{
\begin{tabular}{l|ccccc}
    \hline
    %& \multicolumn{5}{c}{} \\\hline
     & F1@10 & F1@25 & F1@50 & Edit & MoF \\
    \hline
    Unsupervised & 40.8 & 36.2 & 28.1 & 32.4 & 62.5 \\
    $\text{ICC}_\text{2}$ & 51.3 & 46.6 & 36.5 & 44.7 & 61.3 \\
    $\text{ICC}_\text{3}$ & 52.5 & 47.2 & 36.5 & 45.4 & 62.1 \\
    $\text{ICC}_\text{4}$ & 52.6 & 47.7 & 38.1 & 46.7 & 61.3 \\
    \hline
    \end{tabular}
}\end{center}
\caption{\textbf{Improvement in representation} on 50salads for 5\% labelled data with more iterations of ICC. Note: Representation is evaluated with 100\% data with simple Linear Classifier as discussed in section 4.4.}
\label{tab:50Salads_reprentation_learning}
\end{table}

\begin{figure*}
    \centering
    \includegraphics[width=0.9\linewidth]{images/50salads_segmentation_output.pdf}
    \caption{A qualitative example taken from of 50salads, showing progressive improvement in segmentation results with number of iterations of ICC. Some segments becomes more aligned to ground truth(GT) leading improved MoF and F1@50 scores.}
    \label{fig:50salads_icc_example}
\end{figure*}

\begin{table*}[t]
\begin{center}
\small{
% \begin{tabular}{p{0.7cm}|p{1.5cm}|p{0.5cm}p{0.5cm}p{0.5cm}p{0.5cm}p{0.6cm} | p{0.5cm}p{0.5cm}p{0.5cm}p{0.5cm}p{0.6cm} | p{0.5cm}p{0.6cm}p{0.6cm}p{0.5cm}p{0.6cm}}
\begin{tabular}{c|c|ccccc|ccccc|ccccc}
\hline
& & \multicolumn{5}{c|}{Breakfast} & \multicolumn{5}{c|}{50Salads} & \multicolumn{5}{c}{GTEA} \\
\hline
\%$D_L$ &\textbf{Method} & \multicolumn{3}{c}{$F1@\{10,25,50\}$} & Edit & MoF & \multicolumn{3}{c}{$F1@\{10,25,50\}$} & Edit & MoF & \multicolumn{3}{c}{$F1@\{10,25,50\}$} & Edit & MoF \\
\hline
\multirow{4}{*}{\textbf{$\approx$10}} & $\text{ICC}_1$ \rowcolor{LightOrange} 
                    & 57.0 & 51.9 & 36.3 & 56.3 & 65.7 
                    & 51.1 & 45.6 & 34.5 & 42.8 & 65.3 
                    & 82.2 & 78.9 & 63.8 & 75.6 & 72.2\\
& $\text{ICC}_2$ \rowcolor{LightOrange} 
                    & 60.0 & 54.5 & 38.8 & 59.5 & 66.7
                    & 56.5 & 51.6 & 39.2 & 48.9 & 67.1 
                    & 83.4 & 80.1 & 64.2 & 75.9 & 72.9 \\
& $\text{ICC}_3$ \rowcolor{LightOrange} 
                    & 62.3 & 56.5 & 40.4 & 60.6 & 67.8 
                    & 60.7 & 56.9 & 45.0 & 52.4 & 68.2
                    & 83.5 & 80.8 & 64.5 & 76.3 & 73.1\\
& $\text{ICC}_4$ \rowcolor{LightOrange} 
                    & 64.6 & 59.0 & 42.2 & 61.9 & 68.8
                    & 67.3 & 64.9 & 49.2 & 56.9 & 68.6 
                    & 83.7 & 81.9 & 66.6 & 76.4 & 73.3 \\
                    
\cline{2-17}
& Gain & 7.6 & 7.1 & 5.9 & 5.6 & 3.1 
       & 16.2 & 19.3 & 14.7 & 14.1 & 3.3 
       & 1.5 & 3.0 & 2.8 & 0.8 & 1.1 \\
\hline
\multirow{4}{*}{\textbf{100}} & $\text{ICC}_1$ 
\rowcolor{LightGreen} 
                & 68.1 & 63.5 & 49.4 & 66.5 & 72.2
                & 72.2 & 69.1 & 59.9 & 62.2 & 79.7 
                & 89.9 & 88.1 & 79.2 & 84.7 & 80.9\\
& $\text{ICC}_2$ \rowcolor{LightGreen} 
                & 71.5 & 67.6 & 54.5 & 68.0 & 74.9 
                & 78.4 & 75.8 & 68.5 & 71.1 & 82.9
                & 89.9 & 88.3 & 75.6 & 85.3 & 80.9\\
& $\text{ICC}_3$ \rowcolor{LightGreen}
                & 71.9 & 68.2 & 54.9 & 68.5 & 75.0 
                & 81.3 & 79.3 & 71.9 & 74.4 & 84.5
                & 90.1 & 88.3 & 78.3 & 86.8 & 81.0 \\
& $\text{ICC}_4$ \rowcolor{LightGreen}
                & 72.4 & 68.5 & 54.9 & 68.6 & 75.2
                & 83.8 & 82.0 & 74.3 & 76.1 & 85.0 
                & 91.4 & 89.1 & 80.5 & 87.8 & 82.0 \\
\cline{2-17}
& Gain & 4.3 & 5.0 & 5.5 & 2.1 & 3.0 
       & 11.6 & 12.9 & 14.4 & 13.9 & 5.3
       & 1.5 & 1.0 & 1.3 & 3.1 & 1.1 \\
\hline

\end{tabular}
}\end{center}
\caption{Quantitative evaluation of progressive semi-supervised improvement with more iterations of \textbf{ICC with $\approx$ 10\% and 100\% labelled data}.}
\label{tab:icc_10_100}
\end{table*}

\subsubsection{Unsupervised Clusterring vs Representation Learning:} We also report the linear classification scores for the publicly available checkpoints of unsupervised segmentation CTE-MLP~\cite{unsupervised-kukleva2019unsupervised} (which pre-trains representation for predicting the absolute temporal positions of features) in Table~\ref{tab:unsupervised_sota}. Surprisingly, linear classifier accuracy of the representation from there unsupervised work is below the input IDT features baseline for 50Salads. This is likely due to there assumptions made of embedding the absolute temporal positions to the features before clustering, when positions of actions within 50salads is widely different in different videos of the dataset. However, we note, as discussed in related work section 2 of the main paper, unsupervised works like ~\cite{unsupervised-kukleva2019unsupervised, unsupervised-li2021action, sarfraz2021temporally} is based on development of clustering algorithms, with requirement of higher order viterbi algorithm. Using the same representation with two different clustering algorithm, the segmentation results can vary widely. For example, IDT features of 50salads has MoF of 29.7\% with Viterbi+Kmeans and 66.5\% with TWFINCH~\cite{sarfraz2021temporally} as shown in ~\cite{sarfraz2021temporally}. So our unsupervised representation learning step is not directly comparable to unsupervised clustering(segmentation) algorithms. They can only evaluate there unsupervised clusters based on Hungarian Matching to ground truth labels (\ie no classifier is trained, only creates segmentation without labelling and  temporal action segmentation requires to jointly segment and classify all actions) and therefore they cannot be compared to fully-semi-weakly supervised works.

\subsection{ICC without unsupervised step.}\label{subsec:icc_wo_pretraining}

In table~\ref{tab:wo_unsuper_pretrain}, we show results of our ICC without the initial \textit{``unsupervised representation learning''} as \textbf{``ICC-Wo-Unsupervised''}. This essentially means that the $2^{nd}$ row of table~\ref{tab:wo_unsuper_pretrain} represents the scenario in which from our ICC algorithm we remove the $1^{st}$ contrast step that is learned with cluster labels. The improvement in scores over supervised setup is quite low compared to the full-ICC \textit{with} the unsupervised pre-training, shown as \textbf{``ICC-With-Unsupervised''}. This verifies the importance of our unsupervised learning step in ICC.

\subsection{Iterative progression of ICC results} \label{subsec:icc_progression_results}

We discuss our detailed semi-supervised algorithm in section 5 of our main paper and provide visualization of the algorithm in Figure 4. We summarize it with an Algorithm~\ref{algo:contrast_classify} in supplementary.

\begin{table*}[t!]

\begin{center}
\small{
\begin{tabular}{l|l|ccccc}
    \hline
     Dataset & ICC(Num Videos) & F1@10 & F1@25 & F1@50 & Edit & MoF \\
    \hline
    \multirow{4}{*}{Breakfast}  & \rowcolor{LightCyan} $\text{ICC}_1$ ($\approx 63$ Videos) & 54.5 $\pm$  1.2 & 48.7 $\pm$ 1.1  & 33.3 $\pm$ 1.1 & 54.6 $\pm$ 0.9 & 64.2 $\pm$  1.3 \\
    % 2.262211308	2.263271968	2.07595761	1.41788575	2.382771496
    & \rowcolor{LightCyan} $\text{ICC}_4$ ($\approx 63$ videos) &  60.2 $\pm$ 1.5 & 53.5 $\pm$ 1.3 & 35.6 $\pm$ 0.9 & 56.6 $\pm$ 1.2 & 65.3 $\pm$ 1.8 \\
    \cline{2-7}
    & \rowcolor{LightCyan}  $\text{ICC}_1$ ($\approx 120$ Videos) & 57.0 $\pm$ 1.9 & 51.9 $\pm$  2.1 & 36.3 $\pm$ 1.3 & 56.3 $\pm$ 1.2 & 65.7 $\pm$ 1.9 \\
    % 2.262211308	2.263271968	2.07595761	1.41788575	2.382771496
    & \rowcolor{LightCyan}  $\text{ICC}_4$ ($\approx 120$ Videos) &  64.6 $\pm$ 2.1 & 59.0 $\pm$ 1.9 & 42.2 $\pm$ 2.5 & 61.9 $\pm$ 2.2 & 68.8 $\pm$ 1.3 \\\hline
    %& \multicolumn{5}{c}{} \\\hline

    % Base I3D  & 12.2 & 7.9 & 4.0 & 8.4 & 55.0 & 4.9 & 2.5 & 0.9 & 5.3 & 30.2 \\
    % 1.908297671	1.537530488	1.142978565	1.60623784	2.340939982
    \multirow{4}{*}{50salads} &  \rowcolor{LightPink} $\text{ICC}_1$ (3 Videos) & 41.3 $\pm$ 1.9 & 37.2 $\pm$ 1.5 & 27.8 $\pm$ 1.1 & 35.4 $\pm$ 1.6 & 57.3 $\pm$ 2.3 \\
    % 2.262211308	2.263271968	2.07595761	1.41788575	2.382771496
    & \rowcolor{LightPink}  $\text{ICC}_4$ (3 videos) &  52.9 $\pm$ 2.2 & 49.0 $\pm$ 2.2 & 36.6 $\pm$ 2.0 & 45.6 $\pm$ 1.4 & 61.3 $\pm$ 2.3 \\
    \cline{2-7}
    & \rowcolor{LightPink} $\text{ICC}_1$ (5 Videos) & 51.1 $\pm$  2.1 & 45.6 $\pm$  1.3 & 34.5 $\pm$ 1.7 & 42.8 $\pm$ 1.1 & 65.3 $\pm$  0.8 \\
    % 1.695824939	2.480806321	1.873250295	1.057081937	0.819918695
    & \rowcolor{LightPink} $\text{ICC}_4$  (5 videos) & 67.3 $\pm$ 1.8 & 64.9 $\pm$ 2.5 & 49.2 $\pm$ 1.8 & 56.9 $\pm$ 2.1 & 68.6 $\pm$  0.7 \\
    \hline
    
    \end{tabular}
}\end{center}
\caption{\textbf{Mean and standard deviation for 5 different selections} of 5\% and 10\% labelled videos from Breakfast and 50salads. For each metric we report the results in the format $mean \pm std$, \ie the means and the standard deviation for the $5$ runs.}
\label{tab:50Salads_error_bars_5}
\end{table*}

\subsubsection{Improvement after \textit{Contrast} step}
In Table \ref{tab:50Salads_reprentation_learning} we show the improvement in representation after each \textit{contrast} step. Due to the usage of better pseudo-labels obtained from the preceding \textit{classify} step, the following contrast step results in better representations as more iteration is performed. For 5\% labelled videos of the 50salads dataset, we can see that there is a clear improvement in F1 and Edit scores as more iterations are performed. Note that the evaluation of the learned representation is same as described in subsection 4.4 of the main text.

\subsubsection{Improvement after \textit{Classify} step}
In Table 2 of main text, we showed the progressive improvement in performance for 5\% labelled videos, evaluated after the \textit{classify} step of each ICC iteration. In Table~\ref{tab:icc_10_100} we show the same progressive improvements for 10\% and 100\% labelled videos. The evaluation is done after the \textit{classify} step of each iteration of the algorithm.Our ICC raises overall scores on all datasets, with stronger improvements in F1 and Edit scores. 

\textbf{Qualitative visualization of segmentation} We show in Figure~\ref{fig:50salads_icc_example} an example segmentation results from 50salads dataset of how the segmentation results improves (becomes more aligned to GT shown with increase in MoF and F1@50) with more iterations of ICC.   % compared to MF scores. We can intuitively justify 
% Intuitively, we believe that %this is due to the fact that 
% after fine-tuning, % on our labelled data, we improve our 
%contrastive feature learning by 
% the pseudo labels for contrastive feature learning, while not as accurate as true labels, can  %predicted, which smoothens the 
% smooth the segmentation results. 
\subsection{Standard deviations in results} \label{subsec:mean_std_results} We show our standard deviations in results for 50Salads, Breakfast dataset for variations in labelled data used in Table~\ref{tab:50Salads_error_bars_5}. We show the variation in results for $\text{ICC}_1$ and $\text{ICC}_4$ when we take 5 different random selections of 5\%, 10\% labelled videos in Breakfast and 50Salads from corresponding training splits. We report the mean and standard deviation for different choices as $mean \pm std$ format.

\bibliographystyle{aaai22.bst}
\bibliography{aaai22.bib}